\documentclass[sigconf]{acmart}

\AtBeginDocument{%
  }

\setcopyright{acmlicensed}
\copyrightyear{2024}
\acmYear{2024}
\setcopyright{acmlicensed}\acmConference[CIKM '24]{Proceedings of the 33rd ACM International Conference on Information and Knowledge Management}{October 21--25, 2024}{Boise, ID, USA}
\acmBooktitle{Proceedings of the 33rd ACM International Conference on Information and Knowledge Management (CIKM '24), October 21--25, 2024, Boise, ID, USA}
\acmDOI{10.1145/3627673.3679801}
\acmISBN{979-8-4007-0436-9/24/10}

\usepackage{algorithm}
\usepackage{tikz}
\usepackage{multirow}%
\usepackage{amsmath,amsfonts}%
\usepackage{amsthm}%
\usepackage{mathrsfs}%
\usepackage[title]{appendix}%
\usepackage{xcolor}%
\usepackage{textcomp}%
\usepackage{manyfoot}%
\usepackage{booktabs}%
\usepackage{algorithm}%
\usepackage{algorithmicx}%
\usepackage{algpseudocode}%
\usepackage{listings}%
\usepackage{tikz,bm}
\usepackage{enumitem}
\usepackage{xspace}
\usepackage[algo2e,ruled,linesnumbered]{algorithm2e}
\usepackage{lipsum}
\usepackage{subcaption}
\usepackage{bbding}
\usepackage{comment}
\usepackage{newfloat}
\usepackage{listings}
\DeclareCaptionStyle{ruled}{labelfont=normalfont,labelsep=colon,strut=off} 
\floatstyle{ruled}
\newfloat{listing}{tb}{lst}{}
\floatname{listing}{Listing}

\def\etc{etc.\@\xspace}
\def\ie{{\it i.e.},\@\xspace}
\newcommand{\Arrow}[1]{%
\parbox{#1}{\tikz{\draw[->](0,0)--(#1,0);}}
}

\begin{document}

\title{Adaptive Cascading Network for Continual Test-Time Adaptation}

\author{Kien X. Nguyen}
\authornote{Both authors contributed equally to this research.}
\orcid{0009-0002-6470-7278}
\affiliation{%
  \department{Department of Computer and Information Sciences}
  \institution{University of Delaware}
  \city{Newark}
  \state{Delaware}
  \country{USA}
}
\email{kxnguyen@udel.edu}

\author{Fengchun Qiao}
\authornotemark[1]
\orcid{0000-0003-2714-2036}
\affiliation{%
  \department{Department of Computer and Information Sciences}
  \institution{University of Delaware}
  \city{Newark}
  \state{Delaware}
  \country{USA}
}
\email{fengchun@udel.edu}

\author{Xi Peng}
\orcid{0000-0002-7772-001X}
\affiliation{%
  \department{Department of Computer and Information Sciences}
  \institution{University of Delaware}
  \city{Newark}
  \state{Delaware}
  \country{USA}
}
\email{xipeng@udel.edu}


\begin{abstract}
We study the problem of continual test-time adaption where the goal is to adapt a source pre-trained model to a sequence of unlabelled target domains at test time. Existing methods on test-time training suffer from several limitations: 
(1) Mismatch between the feature extractor and classifier; (2) Interference between the main and self-supervised tasks; (3) Lack of the ability to quickly adapt to the current distribution. 
In light of these challenges, we propose a cascading paradigm that simultaneously updates the feature extractor and classifier at test time, mitigating the mismatch between them and enabling long-term model adaptation. 
The pre-training of our model is structured within a meta-learning framework, thereby minimizing the interference between the main and self-supervised tasks and encouraging fast adaptation in the presence of limited unlabelled data.
Additionally, we introduce innovative evaluation metrics, \textit{average accuracy} and \textit{forward transfer}, to effectively measure the model's adaptation capabilities in dynamic, real-world scenarios.
Extensive experiments and ablation studies demonstrate the superiority of our approach in a range of tasks including image classification, text classification, and speech recognition. Our code is publicly available at \href{https://github.com/Nyquixt/CascadeTTA}{\color{magenta} https://github.com/Nyquixt/CascadeTTA}.
\end{abstract}




\maketitle

\section{Introduction}
Intelligent systems operating in real-world settings frequently encounter non-stationary data distributions that evolve over time.
For example, self-driving cars would face dynamically changing environments due to weather, lighting conditions, geographic locations, \etc
This variability poses significant challenges to machine learning models as a highly accurate model on training data may fail catastrophically on shifted distributions \cite{goodfellow2014adv,qiao2021uncertainty,qiao2023topology,peng2022out,li2023data}.
These challenges necessitate model retraining to assimilate new distributions, also known as continual learning \cite{mccloskey1989catastrophic}.
However, most existing continual learning approaches are designed primarily for fully labeled datasets, which is prohibitively expensive.

\begin{figure}
    \centering
    \includegraphics[width=0.8\linewidth]{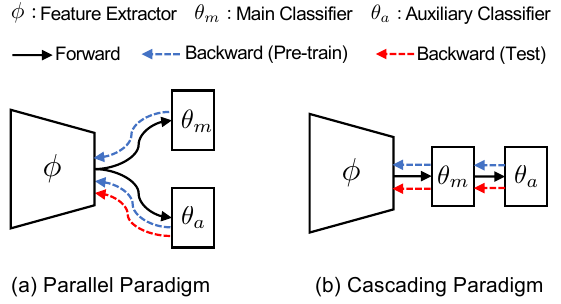}
     \caption{ (a) Parallel paradigm for test-time training \cite{sun2019test}, and (b) our proposed cascading paradigm for continual test-time training.
     The proposed cascading paradigm efficiently mitigates the mismatch between the feature extractor and main classifier, enabling long-term model adaptation.
     }
    \label{fig:models}
    \vspace{-10pt}
\end{figure}

Test-time adaptation \cite{sun2019test,wang2021tent,prabhudesai2023test,qiao2024ensemble} is a promising approach to adapt a pre-trained model to unlabelled target data without access to training data.
The essence of these methods lies in the incorporation of a self-supervised learning (SSL) task, {\it e.g.}, predicting image rotation, minimizing the entropy loss, etc.
Recently, Wang et al.~\cite{Wang2022ContinualTD} has extended the problem setting to continual test-time adaptation, where there is a sequence of target distributions at test time, to better reflect the real-world scenario.
However, recent work Liu et al.~\cite{liu2021ttt} shows that the unconstrained model update from the SSL task may interfere with the main task. This interference results in the accumulation of prediction errors and a gradual deviation from the model's true predictive mechanism, preventing long-term model adaptation.
Moreover, existing methods lack the ability of quickly adapting  ({\it e.g.}, adapting with few gradient steps) the model to the current distribution and demand statistically sufficient data ({\it e.g.} large batch size).
This issue is crucial for continual and non-stationary settings when test data arrive in an online manner with small batches.
In addition, test-time training methods \cite{sun2019test,liu2021ttt,bartler2022mt3} typically employ a parallel paradigm (Fig.~\ref{fig:models} (a)) with an auxiliary classifier for the SSL task, thus only enabling feature update while keeping the main classifier fixed at test time. 
Continually updating the model in such a manner would lead to a growing misalignment between the feature extractor and the main classifier, leading to deteriorated test accuracy.

To address the issues, we propose a cascading paradigm for continual test-time adaptation. 
In contrast to previous parallel paradigm which only enables feature update, the proposed cascade paradigm (Fig.~\ref{fig:models} (b)) synchronously modulates the feature extractor and main classifier at test-time, mitigating the mismatch between them and enables long-term model adaptation.
To optimize the proposed cascading paradigm, we organize the model pre-training in a meta-learning framework, minimizing the interference between the main and SSL tasks and encouraging fast adaptation with limited amounts of unlabelled data.

Given the complexities identified in continual test-time adaptation, we recognize the need for more refined metrics in addition to those tailored for standard test-time adaptation \cite{hendrycks2019benchmarking}. Taking inspiration from the continual learning literature \cite{Vniat2021EfficientCL}, we introduce two new evaluation metrics to the continual test-time adaptation setting: (1) \textit{average accuracy} identifies whether the model has drifted away from the true prediction mechanisms at the end of the adaptation process, and (2) \textit{forward transfer} evaluates the model's capability to leverage knowledge from the past domains to adapt to the current one.
To summarize, our contribution is multi-fold:
\begin{itemize}
    \item We propose a cascading paradigm tailored for continual test-time adaptation to efficiently eliminate the mismatch between the feature extractor and classifier at test-time, enabling long-term model adaptation.
    \item We organize the model pre-training in a meta-learning framework to align the main and SSL tasks, meanwhile encouraging fast adaptation to target distributions in the presence of limited unlabelled data.
    \item We introduce new evaluation metrics, namely average accuracy and forward transfer, to further understand the model's behavior in long-term adaptation.
    \item Extensive experiments demonstrate the superiority of our approach in a wide scope of tasks including image classification, text classification, and speech recognition.
\end{itemize}
\section{Related Work}

{\bf Test-time Adaptation.}
Several methods have been developed to adapt pre-trained models to test data from shifted distributions without accessing source data. 
Liang et al.~\cite{liang2020we} used information maximization and pseudo-labeling for implicit alignment between target and source domains. 
Sun et al.~\cite{sun2019test} adapts the feature extractor using self-supervised tasks like image rotation prediction. 
Batch normalization methods, including re-estimating target domain normalization statistics \cite{li2016revisiting} and introducing entropy-based updates \cite{wang2021tent}, are also utilized.
However, these methods often lack flexibility for continually changing distributions.
Recently, Wang et al.~\cite{Wang2022ContinualTD} formulated continual test-time adaptation (CoTTA) to address the issue of catastrophic forgetting for non-stationary distributional shifts. Despite its effectiveness, CoTTA updates all parameters of model, degrading adaptation efficiency and risking overfitting to data streams. This problem is also known as batch dependency, or \textit{over-adaptation on previous test batches} \cite{zhao2023pitfalls}.

{\bf Continual Learning.}
The objective is to learn progressively from tasks in sequence without erasing previously gained knowledge \cite{de2021continual}. 
Existing methods can be divided into three categories: regularization-based \cite{kirkpatrick2017overcoming,li2017learning}, memory replay \cite{rebuffi2017icarl,chaudhry2018efficient}, and parameter isolation methods \cite{aljundi2017expert,mallya2018packnet}. 
Different from continual test-time adaptation, they focus on learning new tasks with fully labelled datasets. 
Several methods are proposed to adapt models to a sequence of unlabelled target domains, also known as continual domain adaptation \cite{hoffman2014continuous,wulfmeier2018incremental}. 
Liu et al.~\cite{liu2020learning} suggested a meta-adaptation framework capable of capturing the evolving pattern of the target domain. 
Su et al.~\cite{su2020gradient} proposed gradient regularized contrastive learning to learn discriminative and domain-invariant representations simultaneously.
Lao et al.~\cite{lao2021two} introduced a modularized two-stream system which can handle both task and domain shifts.
However, these methods rely on the co-existence of both the source and target domains, and cannot be applied directly in our problem.
Lifelong domain adaptation \cite{rostami2021lifelong,huang2022lifelong} lifts this restriction but necessitates estimating internal source distribution for adaptation.

{\bf Meta-learning.}
Meta-learning \cite{schmidhuber1987evolutionary} is a long-standing topic on learning models to generalize over a distribution of tasks. 
Model-Agnostic Meta-Learning (MAML) by Finn et al.~\cite{finn2017model} was proposed to adapt the model to new tasks within a few gradient steps.
The key idea is to learn a good initialization from which the model is able to be quickly adapted to new tasks with few-shot examples.
Several approaches~\cite{qiao2020learning,ma2021smil} have been proposed to use meta-learning to address distributional shifts. Li et al.~\cite{li2018learning} suggested an episodic training paradigm to improve models' generalization capability.
Balaji et al.~\cite{balaji2018metareg} came up with MetaReg that meta-learned a regularization function that can generalize to new domains. 
Dou et al.~\cite{dou2019domain} proposed to incorporate global and local constraints to learn semantic feature spaces in a modified MAML framework. 
MT3 \cite{bartler2022mt3} was formulated to incorporate meta-learning into test-time adaptation by learning task-specific model parameters for different tasks. 
However, MT3 requires the meta-model to be accessed at test time which makes continual adaptation for a single model unsuitable.
\section{Problem Formulation}\label{sec:problem}

The problem of continual test-time adaptation is defined by a pair of random variables $(X, Y)$ over instances $x \in \mathcal{X} \subseteq \mathbb{R}^{d}$ and corresponding labels $y \in \mathcal{Y}$, where $(X, Y)$ follows an unknown joint distribution $\mathbb{P}(X, Y)$.
At training, we have access to a single source domain $\mathcal{S}$ with labelled dataset $\mathcal{D}_{\mathcal{S}} = \{\boldsymbol{x}^i_s, \boldsymbol{y}^i_s\}^{n_s}_{i=1}$, where $\mathcal{D}_{\mathcal{S}} \sim \mathbb{P}_{\mathcal{S}}(X, Y)$. At test time, we are presented with a sequence of target domains $\mathcal{T} = \left\{\mathcal{T}_{1}, \mathcal{T}_{2}, \cdots, \mathcal{T}_{N}\right\}$ where for each $\mathcal{T}_i$ we only have access to an unlabelled dataset $\mathcal{D}_{\mathcal{T}_i} = \left\{\boldsymbol{x}^k_{t_i}\right\}^{n_{t_i}}_{k=1}$ and $\boldsymbol{x}_{t_i}$ arrive in an online manner with small batches while $\mathcal{D}_{\mathcal{S}}$ is not available:
\begin{equation*}
\boldsymbol{x}_{t_i} \stackrel{i.i.d.}{\sim} \mathcal{D}_{\mathcal{T}_{i}} \text { where } \mathcal{D}_{\mathcal{T}_{1}} \rightarrow \mathcal{D}_{\mathcal{T}_{2}}, \cdots, \rightarrow \mathcal{D}_{\mathcal{T}_{N}} \sim \mathbb{P}_{\mathcal{T}}.
\end{equation*}

The objective is to learn a predictor $f_{\psi}$ : $\mathcal{X} \to \mathcal{Y}$ to predict labels $\{\boldsymbol{y}^k_{t_i}\}^{n_{t_i}}_{k=1}$ for each $\mathcal{T}_{i}$, where $\psi$ are learnable model parameters.
Typically, $f_{\psi}$ is decomposed into a feature extractor $h_{\phi}: \mathcal{X} \rightarrow \mathcal{Z} \subset \mathbb{R}^{p}$ and a classifier $g_{\theta}: \mathcal{Z} \rightarrow \mathcal{Y}$, i.e., $f_{\psi}=g_{\theta} \circ h_{\phi}$, where $\psi=(\phi, \theta)$.

\subsection*{Evaluation Metrics}
The first evaluation metric is the standard mean online error, which measures the average classification error across all batches during adaptation, denoted as $\mathcal{E}(\psi)$.
In addition, motivated by continual learning \cite{Vniat2021EfficientCL}, we propose to evaluate the model in terms of average accuracy and forward transfer. Let $\mathcal{R}_{\psi}(\mathcal{T}_i|\mathcal{T}_j)$ denote the test accuracy on domain $\mathcal{T}_i$ after observing $\mathcal{T}_j$. We measure the average accuracy $\mathcal{A}(\psi)$ on all domains at the end of adaptation:
\begin{equation}
\mathcal{A}(\psi) = \frac{1}{N} \sum_{t=1}^N \mathcal{R}_{\psi}(\mathcal{T}_t|\mathcal{T}_{1:N}).
\label{eq:avg-acc}
\end{equation}
Second, we evaluate forward transfer by measuring the accuracy difference between a model that was updated through the sequence of past domains and a model that was only updated by the last domain:
\begin{equation}
\mathcal{F}(\psi) = \frac{1}{N-1} \sum_{t=2}^{N} \mathcal{R}_{\psi}(\mathcal{T}_t|\mathcal{T}_{1:t}) - \mathcal{R}_{\psi}(\mathcal{T}_t|\mathcal{T}_t).\label{eq:fwt}
\end{equation}
The higher the better for two metrics. A high $\mathcal{A}(\psi)$ denotes that the model does not deviate from the true prediction mechanism; a positive $\mathcal{F}(\psi)$ indicates the model's ability to leverage knowledge from previous domains.
\section{Method}
We introduce a cascading paradigm, as illustrated in Fig.~\ref{fig:models} (b), tailored for continual test-time adaptation.
During training, we update the model on the source domain through a main and a self-supervised learning (SSL) task.
At test time, the feature extractor and classifier are concurrently updated through the SSL task.
This simultaneous update diminishes discrepancies between them, paving the way for sustained model adaptation to target domains.
While a straightforward approach to achieve the cascading paradigm might involve pre-training the model on both the main and SSL tasks using multi-task learning, recent research by~\cite{liu2021ttt} indicates potential pitfalls: in multi-task learning, unrestricted updates from the SSL task can inadvertently conflict with the main task, potentially undermining rather than enhancing test accuracy. Furthermore, multi-task learning does not inherently ensure rapid adaptation to target distributions.
To address these challenges, we propose a meta-learning framework during model pre-training. This framework enforces gradient alignment between the two tasks, effectively reducing task interference and improving test accuracy.

\subsection{Cascading Paradigm}

Domain Adaptation \cite{ben2006analysis} is widely used to address distributional shifts without supervision.
However, it requires the co-existence of both the source and target domains.
Test-time training \cite{sun2019test} was proposed to address this issue by leveraging a SSL task for model adaptation.
This method employs a {\bf parallel paradigm} (Fig.~\ref{fig:models} (a)): a main classifier $\theta_m$ for the supervised learning task, an auxiliary classifier $\theta_a$ for the SSL task, and the two classifiers share the same feature extractor $\phi$. All modules are updated in pre-training by multi-task learning, while only the feature extractor is updated at test-time through the SSL task. However, continually updating the model in such manner would gradually increase the discrepancy between the feature extractor and main classifier, accumulating prediction errors and deviating from the true prediction mechanism. 

To address this issue, we propose a {\bf cascading paradigm} (Fig.~\ref{fig:models} (b)) to synchronously modulate the feature extractor and classifier for continual test-time adaptation.
Specifically, we reorganize the architecture in a sequential manner, starting with the feature extractor $\phi$, followed by the main classifier $\theta_m$ and finally the auxiliary classifier $\theta_a$ in that order.
At test time, the self-supervised loss calculated on the output of $\theta_a$ is utilized to synchronously update both $\phi$ and $\theta_m$.

\begin{figure}[t]
    \centering
    \includegraphics[width=\linewidth]{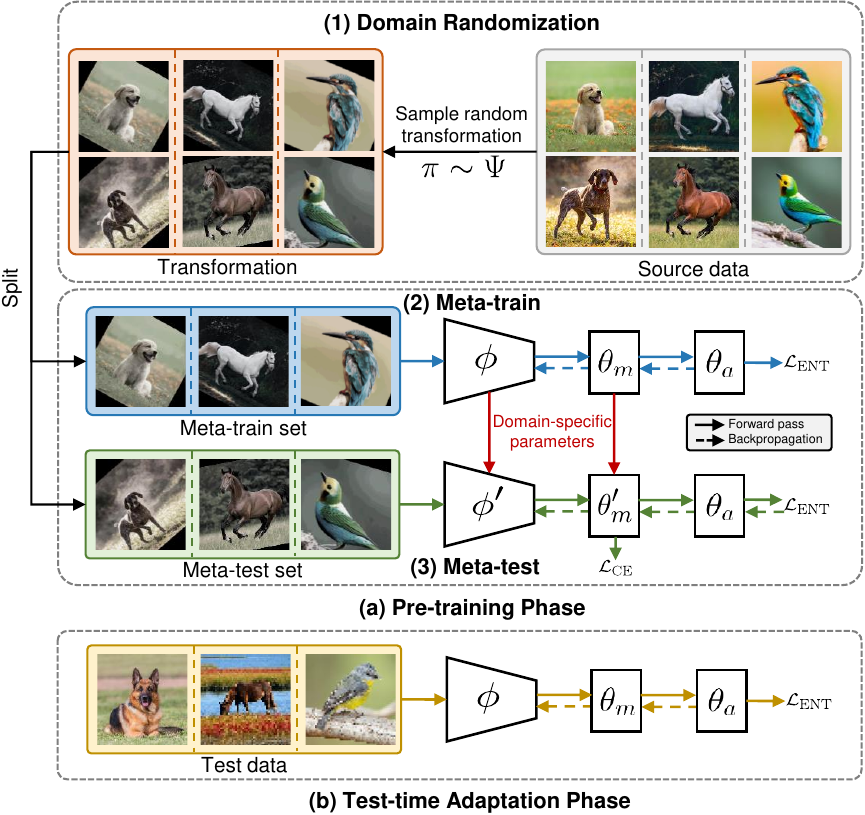}
    \vspace{-5pt}
    \caption{Overview of pre-training and adaptation phases. $\pi \sim \Psi$ denotes a transformation randomly sampled from a predefined pool of transformations. $\mathcal{L}_{\text{ENT}}$: entropy loss, $\mathcal{L}_{\text{CE}}$: cross-entropy loss.}
    \label{fig:overview}
    \vspace{-10pt}
\end{figure}
 
\subsection{Model Pre-training}\label{sec:pre-training}
To optimize the proposed cascading paradigm, a straightforward way is to pre-train the model on both the main and SSL task through multi-task learning. However, recent work by~\cite{liu2021ttt} shows that in multi-task learning, the unconstrained model update from the SSL task may interfere with the main task, deteriorating the test accuracy rather than improving it.
Moreover, multi-task learning cannot guarantee fast adaptation to target distributions.
To address these issues, we propose a meta-learning framework for model pre-training to enforce the gradient alignment between the two tasks, mitigating task interference and enabling fast adaptation with only a single gradient step.
Following \cite{wang2021tent,niu2022efficient}, we employ the entropy loss as the self-supervised loss. 
Our pre-training phase, which includes meta-train and meta-test steps, is structured in a manner that aligns with gradient-based meta-learning approaches like MAML \cite{finn2017model}.


\noindent{\bf Domain Randomization}.
First, we create simulated non-stationary target domains that the model would encounter during test time to support the meta-learning scheme. We employ \textit{domain randomization} \cite{volpi2021continual,qiao2020learning} to generate domain augmentations from the single source $\mathcal{S}$ through a set of transformations $\Psi$. Given $\Psi$ and the source samples $\{(\boldsymbol{x}_s, \boldsymbol{y}_s)\} \sim \mathbb{P}_{\mathcal{S}}(X, Y) $ , we generate augmentations by sampling a specific transformation $\pi \sim \Psi$, and then applying it to the given data, obtaining $\left\{\left(\pi\left(\boldsymbol{x}_{s}\right), \boldsymbol{y}_{s}\right)\right\} \sim  p(\mathcal{S}^+)$. 
For each iteration, we sample a mini-batch $\tau \sim \mathbb{P}_{\mathcal{S}^+}(X, Y)\tau \sim \mathbb{P}_{\mathcal{S}^+}(X, Y)$ representing a random domain.
We then divide $\tau$ into meta-train data $\mathcal{D}_{\tau}^{\text {trn}}$ and meta-validation data $\mathcal{D}_{\tau}^{\text {val}}$ for the meta-learning pre-training phase. 
Domain randomization varies across different modalities. The implementation details are provided in Sec.~\ref{sec: experiments} and Appendix~\ref{app:impl}.

\noindent{\bf Meta-train}.
In the inner loop, to mimic the model adaptation at test-time, we update $\phi$ and $\theta_m$ using the gradients of entropy calculated on the output of $\theta_a$. Specifically, let $\psi=\{\phi, \theta_m\}$,  we update $\psi$ via one step stochastic gradient descent on $\mathcal{D}_{\tau}^{\text {trn}}$ while fixing $\theta_a$:

\begin{equation}
\psi^{\prime} \leftarrow \psi - \alpha \nabla_ {\psi} \mathbb{E}_{\tau \sim \mathbb{P}_{\mathcal{S}^+}}\left[\mathcal{L}_{\text{ENT}}(\psi, \theta_a; \mathcal{D}_{\tau}^{\text {trn}})\right],
\label{eq:fast}
\end{equation}
where $\alpha$ is the learning rate of the inner loop and $\mathcal{L}_{\text{ENT}}$ is the entropy loss. At test time, to avoid overadapting to the data stream and mitigate the risk of catastrophic forgetting, we only update BN parameters of $\phi$. Inspired by the work of \cite{li2016revisiting}, \cite{schneider2020improving}, and \cite{wang2021tent}, we re-estimate the statistical moments and update the affine parameters.

\noindent{\bf Meta-test}.
In the outer loop, we update $\psi$ and $\theta_a$ using the gradient of the meta-loss $\mathcal{L}_{\mathrm{Meta}}$ on $\mathcal{D}_{\tau}^{\text {val}}$:
\begin{equation}
\begin{array}{c}
\{\psi, \theta_a\} \leftarrow \{\psi, \theta_a\} - \beta \nabla_{\{\psi, \theta_a\}}\mathcal{L}_{\mathrm{Meta}}(\psi^{\prime},\theta_a; \mathcal{D}_{\tau}^{\text {val}}),\\ \\
\mathcal{L}_{\mathrm{Meta}} = \mathbb{E}_{\tau \sim \mathbb{P}_{\mathcal{S}^+}}\left[\mathcal{L}_{\mathrm{CE}}\left(\psi^{\prime}; \mathcal{D}_{\tau}^{\text {val}}\right)
+ \lambda\mathcal{L}_{\text{ENT}}\left(\psi^{\prime}, \theta_a; \mathcal{D}_{\tau}^{\text {val}}\right)\right],
\label{eq:meta}
\end{array}
\end{equation}
where $\beta$ is the learning rate of the outer loop,
$\mathcal{L}_{\mathrm{CE}}$ denotes cross-entropy loss, and $\lambda$ is the balancing coefficient.
Intuitively, the meta-learning framework distills knowledge from unlabelled samples and leverages it to facilitate supervised classification.
Moreover, the meta-learned initialization enables the model to fast adapt to target domains using limited unlabeled samples. 
With a few steps of gradient update from the meta-learned representation, the model can produce accurate predictions on the incoming data stream. 
This capability further allows the model to be reliably adapted long-term without deviating too far from the meta-learned initialization, empirically illustrated in Sec.~\ref{sec:img-classification}.
The pre-training procedure is summarized in Algorithm~\ref{alg:overrall}.

It is worth noting that although Tent and CoTTA do not require special treatment to model pre-training, they require statistically sufficient data for model adaptation.
We empirically show that our approach outperforms CoTTA by $\sim9\%$ in accuracy with the batch size of 16 (see Sec.~\ref{sec:ablation}).

{\setlength{\algomargin}{1.5em}
\begin{algorithm2e}[t]
	\caption{Pre-training for Cascading Paradigm.}
	\label{alg:overrall}
	\KwIn
	{Source domain $\mathcal{S}$, transformations $\Psi$}
	\KwOut{Learned $\psi$ and $\theta_a$}
	\While{not converged}{
	     Sample data $\{(\boldsymbol{x}_s, \boldsymbol{y}_s)\} \sim \mathbb{P}_{\mathcal{S}}(X, Y)$ \;
	     Sample a transformation $\pi \sim \Psi$ \;
	     Generate augmentations $\left\{\left(\pi\left(\boldsymbol{x}_s\right), \boldsymbol{y}_s\right)\right\}$ \;
	     Split $\left\{\left(\pi\left(\boldsymbol{x}_s\right), \boldsymbol{y}_s\right)\right\}$ into $\{\mathcal{D}_{\tau}^{\text {trn}}, \mathcal{D}_{\tau}^{\text {val}} \}$ \;
	     \textbf{Meta-train}:
		 Get $\psi^{\prime}$ using $\mathcal{D}_{\tau}^{\text {trn}}$ via Eq.~\eqref{eq:fast} \;
        \textbf{Meta-test}: Update $\{\psi, \theta_a\}$ on $\mathcal{D}_{\tau}^{\text {val}}$ via Eq.~\eqref{eq:meta}\;
   }
\end{algorithm2e}
}
\section{Theoretical Analysis}

We first show that the meta-learning framework encourages gradient alignment between the main and self-supervised learning (SSL) tasks. Next, we show this alignment further upper-bounds the generalization error by adopting the $\mathcal{H}$-divergence~\cite{ben2006analysis}.

\noindent\textbf{Theorem 1.} (Gradient alignment between the main and SSL tasks).
Let $\mathcal{L}_{\text{Main}}(\psi)$ be the loss function for the main task and $\mathcal{L}_{\text{SSL}}(\psi, \theta_a)$ be the loss function for the SSL task.
Suppose the model parameters are updated using Eq.~(\ref{eq:meta}).
Then, the gradient of the meta-objective $\mathcal{L}_{\text{Meta}}$ with respect to $(\psi,\theta_a)$ can be decomposed as:
$$(I - \beta \nabla_{(\psi,\theta_a)}^2 \mathcal{L}_{\text{SSL}}(\psi',\theta_a))^\top (\nabla_{\psi} \mathcal{L}_{\text{Main}}(\psi') + \lambda \nabla_{(\psi,\theta_a)} \mathcal{L}_{\text{SSL}}(\psi',\theta_a)),$$
where $I$ is the identity matrix and $\nabla_{(\psi,\theta_a)}^2 \mathcal{L}_{\text{SSL}}(\psi', \theta_a)$ is the Hessian matrix of $\mathcal{L}_{\text{SSL}}$ with respect to $(\psi,\theta_a)$.
If the Hessian matrix is positive semi-definite and the learning rate $\beta$ is sufficiently small (small $\beta$ ensures that the first-order approximation is valid), the gradient of the SSL task does not interfere with the gradient of the main task, but rather provides useful information for adaptation. 


\noindent\textbf{Theorem 2.} (Generalization Upper Bound. Adapted from ~\cite{liu2020learning}).
{\it Assuming $d_{\mathcal{H} \Delta \mathcal{H}}(\mathbb{P}_{\mathcal{T}_i}, \mathbb{P}_{\mathcal{T}_j}) \leq \alpha\lvert t_i - t_j \rvert$ holds with constant $\alpha$ for $t_i, t_j \geq 0$, then for any $\psi$, with probability at least $1 - \delta$ over the sequence of $N$ target domains $\mathcal{T}$:
{\small
\begin{align*}
 \mathbb{E}_t\mathbb{E}_{\mathbb{P}_\mathcal{T}}\mathcal{L}(f_\psi(\boldsymbol{x}), \boldsymbol{y}) \leq \mathbb{E}_{\mathbb{P}_\mathcal{S}}\mathcal{L}(f_\psi(\boldsymbol{x}), \boldsymbol{y}) &+ \frac{1}{N}\sum_{i=1}^N [d_{\mathcal{H} \Delta \mathcal{H}}(\mathbb{P}_\mathcal{S},\mathbb{P}_{\mathcal{T}_i})] \\
    &+ \mathbb{E}_t\lambda_t + O\big(\frac{\alpha}{\delta N}\big),
\end{align*}
}
\noindent where $\lambda_t = \min_\psi [\mathbb{E}_{\mathbb{P}_\mathcal{S}} \mathcal{L}(f_\psi(\boldsymbol{x}), \boldsymbol{y}) + \mathbb{E}_{\mathbb{P}_\mathcal{T}} \mathcal{L}(f_\psi(\boldsymbol{x}), \boldsymbol{y})]$ measures the adaptability from source to target. 
Our method reduces generalization risks on target domains ($\mathbb{E}_{\mathbb{P}_\mathcal{T}}$ in $\lambda_t$) by aligning the self-supervised learning (SSL) and main tasks.}
\section{Experiments}\label{sec: experiments}

We evaluate our approach on three modalities using five benchmark datasets: {\it CIFAR-10-C}, {\it CIFAR-100-C} and {\it Tiny-ImageNet-C} \cite{hendrycks2019benchmarking} for image classification, {\it Amazon Reviews} \cite{chen2012marginalized} for text classification, and {\it Google Commands} \cite{warden2018speech} for speech recognition.
Section~\ref{sec:ablation} includes ablation studies to investigate key components of the proposed cascading paradigm.
We include source code, implementation details, and more experimental results in the supplementary.

\begin{table*}[t]
\centering
\caption{Results (\%) on {\it CIFAR-10/100-C} and {\it Tiny-ImageNet-C} with the highest corruption severity on the instantaneously changing setup. 
Models are pre-trained on the original {\it CIFAR-10/100} and {\it Tiny-ImageNet} and continually adapted to a sequence of corruptions with a batch size of 32 for {\it CIFAR} and 64 for {\it Tiny-ImageNet}.
Our method significantly outperforms other baselines in online error $\mathcal{E}(\psi)$, average accuracy $\mathcal{A}(\psi)$, and forward transfer $\mathcal{F}(\psi)$.}
\label{tab:main-result}
\resizebox{\linewidth}{!}{
\begin{tabular}{l|l|ccccccccccccccc|ccc}
\toprule
\multirow{2}{*}{Dataset} & \multirow{2}{*}{Method} & \multicolumn{15}{r}{$t$ \Arrow{15.7cm}} & \multirow{2}{*}{$\mathcal{E}(\psi)$ $\downarrow$} &
    \multirow{2}{*}{$\mathcal{A}(\psi) \uparrow$} & \multirow{2}{*}{$\mathcal{F}(\psi) \uparrow$} \\
& & gauss & shot & impul & defoc & glass & motn & zoom & snow & frost & fog & brit & contr & elast & pixel & jpeg &\\
\midrule 
\multirow{6}{*}{CIFAR-10-C} & ERM & 73.06 & 67.73 & 73.38 & 23.71 & 64.87 & 34.53 & 27.08 & 33.41 & 47.54 & \textbf{19.49} & \textbf{10.97} & 23.14 & 39.14 & 73.25 & 36.70 & 43.20 & 56.80 & - \\
& AdaBN & 40.20 & 37.65 & 40.63 & 22.46 & 52.14 & 26.03 & 23.01 & 27.10 & 30.20 & 23.86 & 12.82 & 21.13 & 36.37 & 33.56 & 35.31 & 30.83 & 69.17 & - \\
& TTT & 40.61 & 29.31 & 37.04 & 32.68 & 41.92 & 34.13 & 21.29 & 28.10 & 22.47 & 24.25 &	16.22 &	24.45 & 30.01 & 25.97 & \textbf{24.20} & 28.84 & 64.38 & -4.42 \\	
& Tent & 27.83 & 23.08 & 29.89 & 20.60 & 42.48 & 28.79 & 26.81 & 34.92 & 38.45 & 38.88 & 31.71 & 41.08 & 49.51 & 50.70 & 50.34 & 35.67 & 47.69 & -31.11 \\	
& CoTTA & \textbf{25.63} & \textbf{22.46} & \textbf{27.09} & 19.75 & \textbf{35.35} & 23.10 & 21.33 & 25.16 & 24.91 & 29.15 & 19.15 & 33.40 & 33.22 & 29.04 & 28.87 & 26.51 & 66.18 & -7.49 \\
\cmidrule{2-20}
& Ours & 28.64 & 25.36 & 32.38 &	\textbf{15.52} & 40.55 & \textbf{19.55} & \textbf{16.77} & \textbf{20.65} & \textbf{20.78} & 19.81 & 11.34 & \textbf{16.86} & \textbf{28.36} & \textbf{24.16} & 24.70 & \textbf{22.99} & \textbf{77.36} & \textbf{0.58} \\
\midrule 
\multirow{6}{*}{CIFAR-100-C} & ERM & 94.24 & 91.36 & 91.52 & 55.00 & 87.42 & 62.58 & 56.06 & 67.12 & 76.52 & 57.58 & 42.42 & 66.97 & 65.15 & 90.91 & 67.73 & 71.51 & 28.49 & - \\
& AdaBN & 78.33 & 75.91 & 74.55 & 52.27 & 76.21 & 58.18 & 55.00 & 64.70 & 61.67 & 58.64 & 43.64 & 60.61 & 64.09 & 66.06 & 70.30 & 64.01 & 36.12 & - \\
& TTT & 72.97 & 68.18 & 72.86 & 72.93 & 77.54 & 74.06 & 57.36 & 73.42 & 62.37 & 60.33 & 48.48 & 71.81 & 61.77 & 60.55 & 62.00 & 66.44 & 26.49 & -12.64\\
& Tent & 76.52 & 71.21 & 68.18 & 48.79 & 71.52 & 52.88 & 45.15 & 55.61 & \textbf{53.33} & \textbf{52.42} & 41.21 & 54.55 & \textbf{55.00} & \textbf{52.27} & \textbf{59.39} & 57.20 & 38.84 &  -2.30 \\	
& CoTTA & 68.79 & 70.61 & 69.39 & 57.88 & 70.00 & 61.36 & 55.61 & 61.36 & 56.67 & 62.27 & 45.00 & 75.61 & 64.09 & 60.00 & 60.30 & 62.60 & 35.37 & \textbf{1.52} \\
\cmidrule{2-20}
& Ours & \textbf{66.42} & \textbf{64.78} & \textbf{65.65} & \textbf{44.33} & \textbf{67.20} & \textbf{47.89} & \textbf{46.01} & \textbf{54.11} & 54.35 & 53.75 & \textbf{39.85} & \textbf{49.87} & 57.46 & 54.66 & 60.87 & \textbf{55.11} & \textbf{45.23} & 0.19 \\
\midrule 
\multirow{6}{*}{\shortstack[l]{Tiny\\ImageNet-C}} & ERM & 86.06 & 83.94 & 95.00 & 88.64 & 92.27 & 73.79 & 77.27 & 65.45 & 61.21 & 75.45 & 57.42 & 95.45 & 81.97 & 64.55 & \textbf{48.94} & 76.49 & 23.51 & - \\
& AdaBN & 85.84 & 83.43 & 89.35 & 82.70 & 91.00 & 74.67 & 75.98 & 75.23 & 74.43 & 78.94 & 70.90 & 94.44 & 77.76 & 72.01 & 73.80 & 80.03 & 19.97 & - \\
& TTT & 95.29 & 78.44 & \textbf{79.32} & 82.74 & 88.30 & 81.13 & 67.35 & 75.47 & 65.57 & 69.71 & 60.50 & 96.02 & 67.32 & 59.40 & 57.01 & 74.90 & 20.25 & -11.16 \\
& Tent & 85.22 & 82.79 & 88.23 & 84.40 & 91.61 & 80.31 & 82.26 & 83.49 & 84.17 & 86.33 & 83.14 & 96.48 & 89.76 & 88.28 & 89.90 & 86.42 & 7.72 & -12.58 \\
& CoTTA & 83.49 & 80.37 & 87.86 & 79.22 & 89.18 & 69.65 & 71.10 & 70.83 & 69.49 & 74.49 & 64.67 & 93.07 & 73.35 & 65.33 & 68.00 & 76.01 & 24.60 & \textbf{1.29} \\
\cmidrule{2-20}
& Ours & \textbf{74.62} & \textbf{70.12} & 80.72 & \textbf{77.56} & \textbf{83.63} & \textbf{54.13} & \textbf{54.04} & \textbf{59.22} & \textbf{53.69} & \textbf{63.77} & \textbf{49.34} & \textbf{91.72} & \textbf{63.29} & \textbf{53.84} & 49.56 & \textbf{65.28} & \textbf{27.39} & -12.27 \\
\bottomrule
\end{tabular}
}
\end{table*}

\noindent {\bf Baselines.} 
We compare the proposed cascading paradigm to the following baselines:
(1) Empirical Risk Minimization (ERM) \cite{vapnik1998statistical}: is trained on the source domain and directly evaluated on target domains without any update.
(2) Adaptive Batch Normalization (AdaBN) \cite{Li2018AdaptiveBN}: re-estimate normalization statistics on each incoming batch in target domains.
(3) Test-time training (TTT) \cite{sun2019test}: use a self-supervised learning task, \textit{i.e.}, predicting image rotation, to update the feature extractor at test time.
(4) Tent \cite{wang2021tent}: update both normalization statistics and affine parameters by entropy minimization, modified to fit the continual setting by not resetting during the adaptation process.
(5) CoTTA \cite{Wang2022ContinualTD}: continually adapt a source pre-trained model to non-stationary target data by reducing error accumulation and alleviating forgetting.

For fair comparison, we use our pre-trained model as the backbone for all baselines except ERM and TTT, since TTT has its own pre-training strategy. In contrast to previous work that utilizes pre-trained models such as WideResNet-28 \cite{Zagoruyko2016WideRN}, \textit{we train all models from scratch and refrain from using the data augmentations that coincide with the corruptions/domains at test time to avoid test data leakage, accurately measuring the adapting capability of the algorithms.}

\vspace{1em}
\noindent {\bf Metrics.} {\bf 1) Online error.} 
We immediately record the online prediction of each batch after the model was adapted to it.
We calculate the online prediction error, $\mathcal{E}(\psi)$, by averaging the errors of all target domains.
{\bf 2) Continual learning metrics.}
We further evaluate the cascading paradigm using the metrics proposed in Section~\ref{sec:problem}: average accuracy $\mathcal{A}(\psi)$ and forward transfer $\mathcal{F}(\psi)$.

\subsection{Image Classification} \label{sec:img-classification}

For image classification, We validate our method on {\it CIFAR-10/100-C} and {\it Tiny-ImageNet-C}.

{\bf Dataset.} {\it CIFAR-10/100-C} and {\it Tiny-ImageNet-C} \cite{hendrycks2019benchmarking} is a robustness benchmark consisting of fifteen corruptions types with five levels of severity applied to the test set of {\it CIFAR-10/100} \cite{krizhevsky2009learning} and {\it Tiny-ImageNet} \cite{Le2015TinyIV}. The corruptions consist of four main categories: noise, blur, weather, and digital. We show the model performance on the highest severity.

{\bf Setup.} Following \cite{sun2019test}, we use 15 corruptions as target domains.
The model is pre-trained on the original {\it CIFAR-10/100} and {\it Tiny-ImageNet} datasets and continually adapted to a sequence of image corruptions in {\it CIFAR-10/100-C} and {\it Tiny-ImageNet-C}.
We use ResNet-26 \cite{he2016deep} for the first two datasets and ResNet-34 for the third dataset. We append a lightweight 2-layer fully connected network as the auxiliary classifier. During pre-training, the initial learning rate is 0.001 with a linear decay and the number of epochs is 75.
We use AugMix \cite{hendrycks2020augmix} for domain randomization.
At test time, the SGD optimizer with Nesterov momentum \cite{Sutskever2013OnTI} with an online learning rate of 0.001 updates the model. As we wish to simulate the online setting, we use a small batch size of 32 for {\it CIFAR-10/100-C} and 64 for the more challenging benchmark {\it Tiny-ImageNet-C}.

\begin{table}[ht]
\centering
\caption{Results (\%) on {\it CIFAR-10-C} on the gradually changing setup. Models are pre-trained on {\it CIFAR-10} and continually adapted to a sequence of 135 gradually changing  domains with a  batch size of 32.
Our method enables long-term adaptation to target domains while others  fail catastrophically.
}\label{tab:gradual-c10}
\resizebox{0.55\linewidth}{!}{
\begin{tabular}{l|ccc}
\toprule
Metrics & $\mathcal{E}(\psi)$ $\downarrow$ & $\mathcal{A}(\psi) \uparrow$ & $\mathcal{F}(\psi) \uparrow$ \\
\midrule
TTT & 28.39 & 60.29 & -13.94 \\
Tent & 64.26 & 11.39 & -72.83 \\
CoTTA & 41.08 & 26.93 & -54.83 \\
\midrule
Ours & \textbf{17.51} & \textbf{81.19} & \textbf{-2.4} \\
\bottomrule
\end{tabular}}
\end{table}

{\bf Results.}
{\bf 1) Instantaneously changing setup.}
Tab.~\ref{tab:main-result} shows the results of {\it CIFAR-10-C}, {\it CIFAR-100-C}, and {\it Tiny-ImageNet-C} with the highest corruption severity on the standard domain sequence. Our method outperforms other baselines in most corruptions and yields the lowest average error of $22.99\%$, $55.11\%$, and $65.28\%$ on the three benchmarks, respectively.
For forward transfer, our model achieves a $\mathcal{F}(\psi)$ of $0.58\%$ and $0.19\%$ on {\it CIFAR-10/100-C}, showcasing its capability to adapt to current domains by leveraging past knowledge. However, on the more challenging benchmark {\it Tiny-ImageNet-C}, it suffers a relatively high negative $\mathcal{F}(\psi)$ of $-12.27$.
Other baselines exhibit lower performance on online error with negative forward transfer, except for CoTTA on {\it CIFAR-100-C}, and {\it Tiny-ImageNet-C} with a forward transfer of $1.52\%$, and $1.29\%$.
The results underscore our method's efficiency in continual adaptation to various image corruptions.

{\bf 2) Gradually changing setup.} Next, we evaluate our model on the gradually changing setup, which is more relevant to the real-world scenario. In the standard sequence of corruptions, target domains change abruptly, especially in the highest severity. Instead, following~\cite{Wang2022ContinualTD}, we  construct a sequence of target domains such that the evolving pattern of distributional drift is smoother. Specifically, for each corruption $t$  we continuously change its severity level from 1 to 5 and then back to 1 and then change the corruption type, making up 135 target domains:
\begin{align*}
    \underbrace{...2 \rightarrow 1}_{t-1 \text{ and before}} \rightarrow \underbrace{1 \rightarrow 2 \rightarrow ... 5 \rightarrow 4 \rightarrow ... 1}_{\text{corruption } t} \rightarrow \underbrace{1 \rightarrow 2...}_{t+1 \text{ and after}}
\end{align*}

\noindent This setup evaluates the model's capability of long-term adaptation. 
Per Tab.~\ref{tab:gradual-c10},
Tent and CoTTA, with the need of statistically sufficient data, \ie a large batch size, perform rather poorly.
They both suffer from error accumulation over a lengthy sequence of domains and eventually deviate from the true prediction mechanism.
Since CoTTA updates all model parameters, continually adapting to small batches causes the model to overfit to the current batch, preventing it from efficiently adapt to the subsequent ones.
In contrast, our model maintains stability over an extended domain sequence, achieving a substantial $81.19\%$ average accuracy with minimal forward transfer losses. Our approach alleviates batch dependency \cite{zhao2023pitfalls} by successfully adapting over 135 diverse domains.

\subsection{Text Classification}

In this section, we conduct experiments on {\it Amazon Reviews} by~\cite{chen2012marginalized} for text classification.

{\bf Dataset.}
{\it Amazon Reviews} is a widely adopted benchmark in the context of domain adaptation for sentiment classification. It is a collection of product reviews from amazon.com in four product domains:  books, dvds, electronics, and kitchen appliances. Reviews are assigned with binary labels - 0 (or ``negative'') if the rating of the product is up to 3 stars, and 1 (or ``positive'') if the rating is 4 or 5 stars. We use unigrams and bigrams as features resulting in 5000 dimensional representations \cite{ganin2015unsupervised}.
Following~\cite{wei2019eda}, we leverage the four operations for text augmentation: synonym replacement, random insertion, random swap and random deletion. See Appx.~\ref{sec:text_a}  for more details on augmentations.

{\bf Setup.} We pre-train the model on one source domain ``books'', and continually adapt the model to ``dvds'', ``electronics'', and ``kitchen''. 
Similar to~\cite{ganin2015unsupervised}, the extracted features are fed into two FC layers with the size of 50. A softmax layer with the size of two is used to classify the sentiment of reviews into ``positive'' or ``negative''. We append a BN layer to each hidden layer and append a FC layer with the size of 2 as the auxiliary classifier.
We use Adam \cite{kingma2014adam} to optimize the model, and the initial learning rate $\eta_{0}=10^{-4}$. Similar to the image classification task, the batch size is set to 32.

\begin{table}[t]
\centering
\caption{Results (\%) on {\it Amazon Reviews}. Models are pre-trained on ``books'' and continually adapted to ``dvds'' $\rightarrow$ ``electronics'' $\rightarrow$ ``kitchen'' with a batch size of 32.}\label{tab:text}
\resizebox{\linewidth}{!}{
\begin{tabular}{l|ccc|ccc}
    \toprule
   \multirow{2}{*}{Method}  & \multicolumn{3}{c}{$t$ \Arrow{4cm}} &
    \multirow{2}{*}{$\mathcal{E}(\psi)$ $\downarrow$} &
\multirow{2}{*}{$\mathcal{A}(\psi) \uparrow$} & \multirow{2}{*}{$\mathcal{F}(\psi) \uparrow$} \\
    & dvds & electronics & kitchen &  \\
    \midrule
    ERM & 19.78 & 26.50 & \textbf{20.06} & 22.11 & 77.89 & - \\
    AdaBN & 20.47 & 25.03 & 20.99 & 22.17 & 77.83 & - \\
    Tent & 20.87 & 25.26 & 21.85 & 22.66 & 77.26 & -1.08 \\
    \midrule
    Ours & \textbf{19.19} & \textbf{25.00} & 21.01 & \textbf{21.73} & \textbf{78.73} & \textbf{4.82} \\
    \bottomrule
\end{tabular}}
\end{table}

{\bf Results.}
Tab.~\ref{tab:text} shows the results of text classification on {\it Amazon Reviews}. Results show that our method outperforms others across all test domains by a large margin. Specially, our method achieves the lowest online prediction error of $21.73\%$ and the highest average accuracy of $78.73\%$ across all past domains at the end of the adaptation. Notably, our model achieves $4.82\%$ in forward transfer, indicating the model can leverage knowledge from past domains to accelerate the adaptation to the current domain.
Results demonstrates the effectiveness of our method in continually adapting to different text domains.

\subsection{Speech Recognition}
In this section, we perform experiments on {\it Google Commands} by~\cite{warden2018speech} for speech recognition.

\begin{table}[ht]
\centering
\caption{Results (\%) on {\it Google Commands}. Models are pre-trained on the clean audios and continually adapted to ``Amp.'' $\rightarrow$ ``Pit.'' $\rightarrow$ ``Noise'' $\rightarrow$ ``Strech'' $\rightarrow$ ``Shift''.}
\label{tab:speech}
\resizebox{\linewidth}{!}{
\begin{tabular}{@{}l|ccccc|ccc}
	\toprule
	\multirow{2}{*}{Method} & \multicolumn{5}{r}{$t$ \Arrow{4.6cm}} & 
    \multirow{2}{*}{$\mathcal{E}(\psi)$ $\downarrow$} &
    \multirow{2}{*}{$\mathcal{A}(\psi) \uparrow$} & \multirow{2}{*}{$\mathcal{F}(\psi) \uparrow$} \\
	 & Amp. & Pit. & Noise & Stretch & Shift &  \\
	\midrule
	ERM & 36.5 & 26.7 & 27.4 & 25.8 & 31.7 &  29.62 & 70.38 & - \\
	AdaBN & 36.8 & 26.1 & 26.9 & 22.5 & 28.3 & 28.12 & 71.88 & - \\
	Tent & 36.1 & 23.6 &  24.8 & \textbf{20.2} & 28.5 & 26.64 & 72.25 & -23.15 \\
	\midrule
    Ours & \textbf{35.4}  & \textbf{21.2} & \textbf{21.9} & 20.6 & \textbf{27.3} & \textbf{25.28} & \textbf{74.66} & \textbf{-6.04} \\
	\bottomrule
\end{tabular}}
\end{table}

{\bf Dataset.} {\it Google Commands} has 65000 utterances (one second long) from thousands of people. There are 30 different command words in total. There are 56196, 7477, and 6835 examples for training, validation, and test. To simulate domain shift in real-world scenarios, we apply five common corruptions in both time and frequency domains. This creates five test sets that are ``harder'' than training sets, namely amplitude change (Amp.), pitch change (Pit.), background noise (Noise), stretch (Stretch), and time shift (Shift). 
The range of ``amplitude change'' is (0.7,1.1). The maximum scales of ``pitch change'', ``background noise'', and ``stretch'' are 0.2, 0.45, and 0.2, respectively. The maximum shift of ``time shift'' is 8.
See Appx.~\ref{sec:speech_a} for more details on data preprocessing and augmentations.
We use the default hyper-parameters from audiomentations\footnote{\url{https://github.com/iver56/audiomentations}}.  

{\bf Setup.} We pre-train the model on the clean train set, and continually adapt it to ``Amp.'', ``Pit.'', ``Noise'', ``Stretch'', and ``Shift''. We encode each audio into a Mel-spectrogram with the size of $1 \times 32 \times 32$ and feed them to LeNet \cite{Lecun98gradient} as one-channel input. 

{\bf Results.}
Tab.~\ref{tab:speech} shows the results of speech recognition on {\it Google Commands}. As seen, our method outperforms other baselines on all target domains except ``stretch'', indicating its strong adaptation capability on corruptions in both time and frequency domains. In detail, our method outperforms the second best  by $0.7\%$ on ``amplitude change'', $2.4\%$ on ``pitch change'', $2.9\%$ on ``background noise'', and $1.0\%$ on ``time shift'', respectively. Our method suffers significantly less deterioration in forward transfer compared to Tent.

\subsection{Ablation Study}\label{sec:ablation}

We have shown that our method yields significant improvements across different modalities including image, text, and speech. 
In this section, we perform ablation study to investigate the key components of the proposed cascading paradigm and provide {\it main observations} as follows:

\begin{table}[t]
\centering
\caption{Online classification error (\%) on {\it CIFAR-10-C} across different batch sizes on the instantaneously changing setting.
Our method shows consistent performance across different batch sizes, and outperforms Tent and CoTTA by at least $9\%$ with a batch size of 16. We report the mean and standard deviation across all batch sizes.
}\label{tab:insensitive}
\resizebox{0.8\linewidth}{!}{
\begin{tabular}{l|cccc|c}
\toprule
Batch Size& 128 & 64 & 32 & 16 & Mean $\pm$ Std \\
\midrule
Tent & 22.53 & 25.99 & 35.67 & 50.28 & 33.61 $\pm$ 12.4 \\
CoTTA & 21.75 & 22.96 & 26.51 & 33.58 & 26.20 $\pm$ 5.3 \\

\midrule
TTT & 28.15 & 27.02 & 28.84 & 27.15 & 27.79 $\pm$ \textbf{0.7} \\
Ours & 22.26 & 22.32 & 22.99 & 24.63 & \textbf{23.05} $\pm$ 1.1 \\

\bottomrule
\end{tabular}}
\vspace{-9pt}
\end{table}

{\bf Our model is insensitive to batch sizes. } 
In online adaptation, we expect the batch size to be relatively small and varied; an online model should perform well on such conditions. We show the classification error on {\it CIFAR-10-C} across different batch sizes in Tab.~\ref{tab:insensitive}. A smaller standard deviation indicates that a model is less sensitive to the batch size. CoTTA's performance drops by $11.83\%$ when the batch size reduces from 128 to 16 due to overfitting on early batches with insufficient statistical input. Tent, with limited data, shows $50.28\%$ average error for batch size 16. In contrast, our method exhibits the lowest error, remaining consistent across batch sizes with a $0.9\%$ standard deviation, showcasing its insensitivity and online adaptability It is also worth noting that both TTT and our method are insensitive to different batch sizes, possibly due to the usage of the auxiliary classifier.

\begin{figure}[htbp]
  \centering
  \begin{minipage}[b]{0.48\linewidth} 
    \centering
    \includegraphics[width=\textwidth]{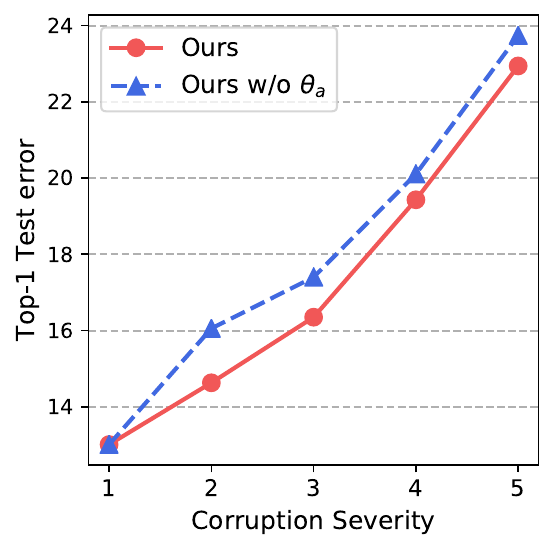}
  \end{minipage}
  \hfill
  \begin{minipage}[b]{0.48\linewidth} 
    \centering
    \includegraphics[width=\textwidth]{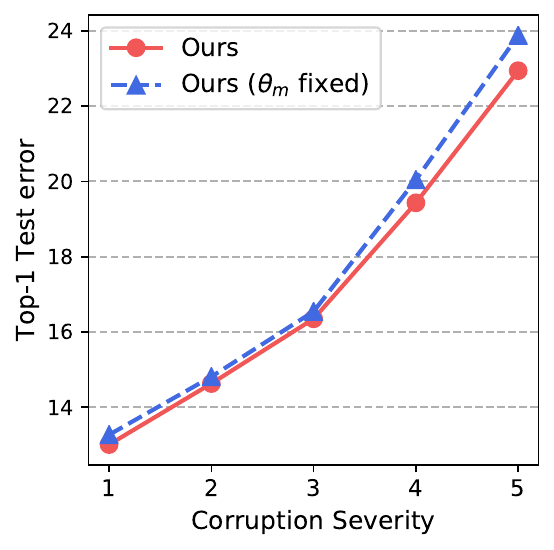}
  \end{minipage}
  \caption{Validation of the cascading paradigm on {\it CIFAR-10-C}. {\bf Left:} Classification error of models w/ and w/o the auxiliary classifier $\theta_a$. 
{\bf Right:} Classification error of models updating and fixing the main classifier $\theta_m$ at test time.}
    \label{fig:cascade}
\end{figure}

{\bf Auxiliary classifier effectively synchronizes modules.} The cascading paradigm allows us to synchronously modulate the feature extractor $\phi$ and main classifier $\theta_m$ through self-supervised learning at test time. To validate its effectiveness, we conduct two ablation studies on \textit{CIFAR-10-C} dataset: 
(1) We discard $\theta_a$ and compute the entropy of the output of $\theta_m$.
In Fig.~\ref{fig:cascade} (left), we observe that $\theta_a$ consistently improves the classification performance across all levels of corruption severity, possibly due to that $\theta_a$ converts the logits into another space with less intervention to the cross-entropy.
(2) We only modulate BN-related parameters and keep $\theta_m$ fixed at test time.
In Fig.~\ref{fig:cascade} (right), we can see that our model outperforms the variant with $\theta_m$ fixed during test, validating the effectiveness of jointly modulating both $\phi$ and $\theta_m$.
We further conduct the ablation study on \textit{Tiny-ImageNet-C}, a more challenging dataset, to confirm the effectiveness of $\theta_a$, as depicted in Tab.~\ref{tab:effect-theta-a}.

\begin{table}
    \centering
    \caption{Ablation study on the effectiveness of $\theta_a$. Online classification error (\%) on {\it Tiny-ImageNet-C} on 5 levels of corruption severity.}
    \resizebox{0.75\linewidth}{!}{
    \begin{tabular}{l|c|c|c|c|c}
    \toprule
         Methods & 1 &  2 &  3 &  4 &  5  \\
    \midrule
         w/o $\theta_a$ & 46.07&  50.46& 56.47&  63.82&  68.09\\
         Ours & \textbf{44.20} & \textbf{48.36} & \textbf{54.14} & \textbf{61.45} & \textbf{65.28}\\
    \bottomrule
    \end{tabular}
    }
    \label{tab:effect-theta-a}
\end{table}

\begin{table}[h!]
\centering
\caption{Ablation study on the effect of meta-learning on {\it CIFAR-10-C} across different batch sizes. 
}\label{tab:ablation-meta}
\resizebox{0.7\linewidth}{!}{
\begin{tabular}{l|c|cccc}
\toprule
Method & Meta & 128 & 64 & 32 & 16 \\
\midrule
TTT & \XSolidBrush & 28.15 & 27.02 & 28.84 & 27.15 \\
TTT & \Checkmark &  24.04 & 24.37 & 27.93 & 28.20 \\
\midrule
Ours & \XSolidBrush & 27.93 & 47.18 & 66.53 & 75.88  \\
Ours & \Checkmark & \textbf{22.26} & \textbf{22.32} & \textbf{22.99} & \textbf{24.63} \\
\bottomrule
\end{tabular}}
\end{table}

{\bf Meta-learning improves model robustness.}
The meta-learning approach aligns gradients between self-supervised and supervised losses and enables effective initialization. This allows the model to rapidly adapt to unlabeled target domains with limited data. To validate its effectiveness, we create a model variant using conventional multi-task learning (MTL). In this variant, we update the model via gradients from the combined entropy and cross-entropy loss: 
\begin{equation*}
\begin{array}{c}
    \{\psi, \theta_a\} \leftarrow \{\psi, \theta_a\} - \alpha\nabla_{\{\psi, \theta_a\}}\mathcal{L}_{\text{MTL}}(\psi,\theta_a;\mathcal{D}_\tau), \\ \\
    \mathcal{L}_{\text{MTL}} = \mathbb{E}_{\tau \sim \mathbb{P}_{\mathcal{S}^+}}\left[ \mathcal{L}_{\mathrm{CE}}\left(\psi; \mathcal{D}_{\tau}\right) + \lambda\mathcal{L}_{\text{ENT}} \left(\psi, \theta_a; \mathcal{D}_{\tau}\right)  \right].
\end{array}
\end{equation*}
Furthermore, we apply meta-learning on TTT and achieve a significant improvement in performance across different batch sizes, as shown in Tab.~\ref{tab:ablation-meta}. 
Similar to Eq.~\ref{eq:fast} and Eq.~\ref{eq:meta} described in Sec.~\ref{sec:pre-training}, we adapt the inner and outer loop update to TTT respectively as:
\begin{equation}
    \phi^{\prime} \leftarrow \phi - \alpha \nabla_ {\phi} \mathbb{E}_{\tau \sim \mathbb{P}_{\mathcal{S}^+}}\left[\mathcal{L}_{\text{R}}(\phi, \theta_a; \mathcal{D}_{\tau}^{\text {trn}})\right],
\end{equation}
and
\begin{equation}
\begin{array}{c}
\{\psi, \theta_a\} \leftarrow \{\psi, \theta_a\} - \beta \nabla_{\{\psi, \theta_a\}}\mathcal{L}_{\mathrm{Meta}}(\phi^{\prime}, \theta_m, \theta_a; \mathcal{D}_{\tau}^{\text {val}}),\\ \\
\mathcal{L}_{\mathrm{Meta}} = \mathbb{E}_{\tau}\left[\mathcal{L}_{\mathrm{CE}}\left(\phi^{\prime}, \theta_m; \mathcal{D}_{\tau}^{\text {val}}\right)
+ \lambda\mathcal{L}_{\text{R}}\left(\phi^{\prime}, \theta_a; \mathcal{D}_{\tau}^{\text {val}}\right)\right],
\end{array}
\end{equation}
where $\mathcal{L}_{\text{R}}$ is the image rotation prediction loss.
Noticeably, it boosts TTT's accuracy with the batch size of 128 by $4.11\%$. Another critical observation is that without meta-learning our cascading paradigm exhibits a poor performance for small batch sizes and has to rely on sufficient data statistics to yield a decent accuracy.
Results demonstrate the effectiveness of meta-learning in adapting to few samples.

\subsection{Uncertainty Quantification}
\begin{figure}[htbp]
  \centering
  \begin{minipage}[b]{0.48\linewidth} 
    \centering
    \includegraphics[width=\textwidth]{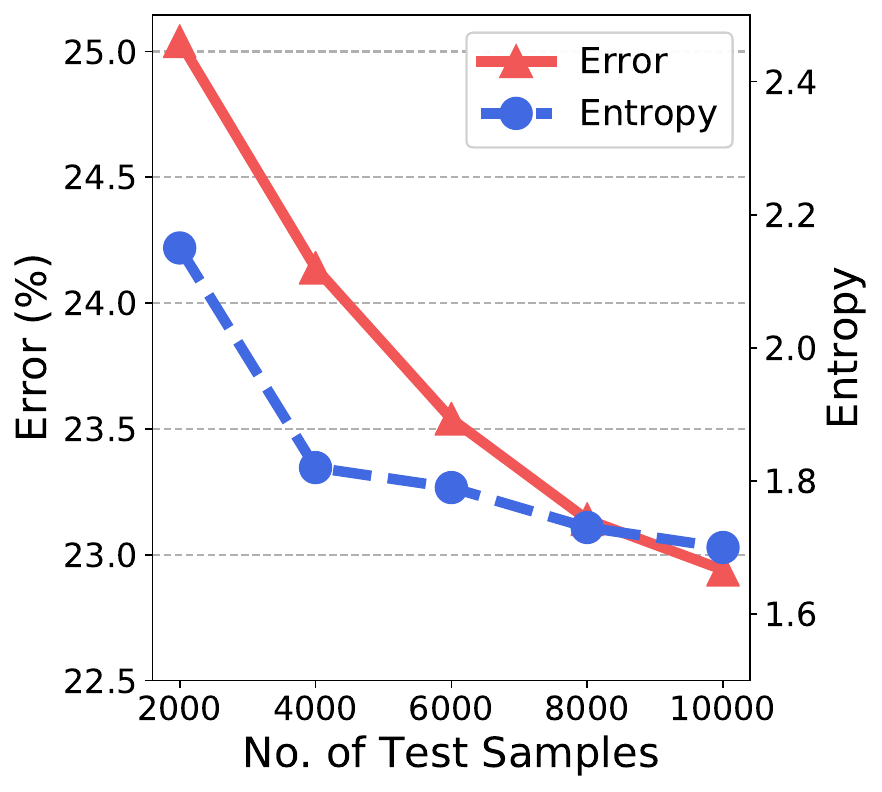}
  \end{minipage}
  \hfill
  \begin{minipage}[b]{0.48\linewidth} 
    \centering
    \includegraphics[width=\textwidth]{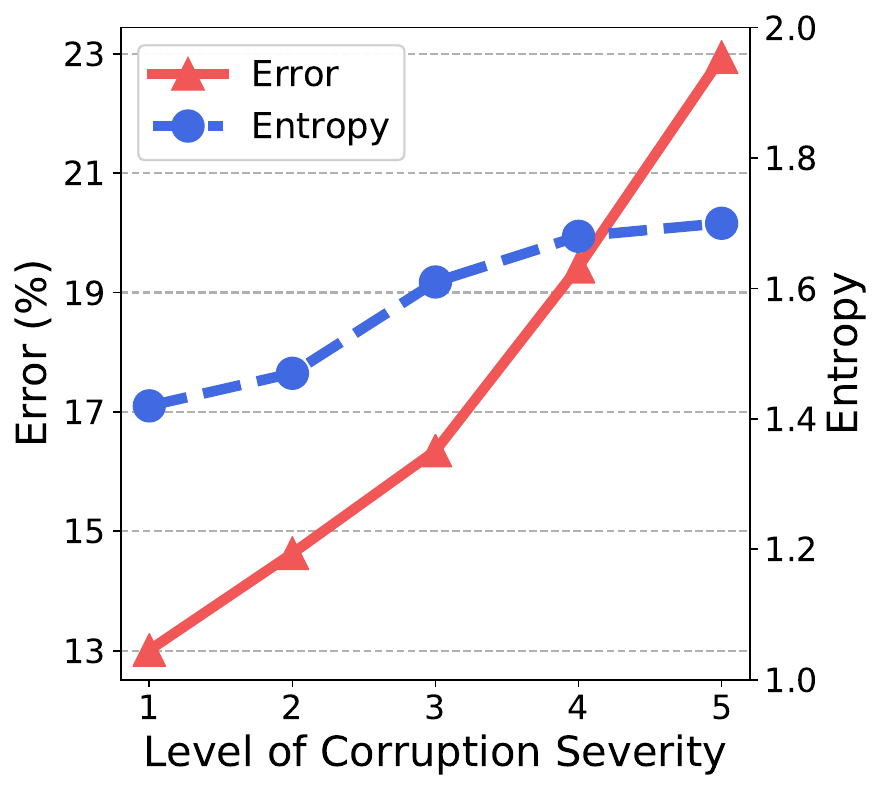}
  \end{minipage}
  \caption{Uncertainty quantification on {\it CIFAR-10-C}. Entropy and error after adapting to different numbers of test samples (\textbf{left}) and different levels of corruption severity (\textbf{right}). Entropy shows consistent trends with the classification error.}
    \label{fig:unc}
    \vspace{-10pt}
\end{figure}

In our approach, we use entropy as the self-supervised loss, which is used to quantify the uncertainty.
To investigate the relationship between uncertainty and classification error, we analyze the evolution of entropy and error. Results are shown in Fig.~\ref{fig:unc}. In Fig.~\ref{fig:unc} (left), we note that both entropy and error decrease when the model adapts to more samples of the target domain. In Fig.~\ref{fig:unc} (right), we can see that both entropy and error increase with the level of severity.
Entropy shows consistent trends with the classification error in both cases.
The results demonstrate that entropy can reflect the model's uncertainty with respect to target domains as well as measure the distributional distance between the source and target domain. 

\section{Conclusion}

We proposed a cascading paradigm for continual test-time adaptation.
In contrast to the parallel paradigm which only enables feature updates, the cascade paradigm synchronously modulates the feature extractor and classifier at test-time, mitigating the mismatch between them and enabling long-term model adaptation.
To minimize the interference between the main and self-supervised tasks, we propose a meta-learning framework for model pre-training to align the gradients between the supervised and self-supervised losses. 
Moreover, the meta-learning framework facilitates fast adaptation to target distributions using limited unlabelled data.
Extensive experiments as well as ablation studies demonstrate the superiority of our approach on a range of tasks including image classification, text classification, and speech recognition.

\begin{acks}
This work is supported by the National Science Foundation through the Faculty Early Career Development Program (NSF CAREER) Award NSF-IIS-2340074 and the Department of Defense under the Defense Established Program to Stimulate Competitive Research (DoD DEPSCoR) Award.
\end{acks}

\vfill\eject

\appendix




\begin{table*}[!htb]
\centering
\caption{Results (\%) on {\it CIFAR-10-C} with corruption level ranging from $1$ to $4$ following the instantaneously changing setup. 
Models are pre-trained on the original {\it CIFAR-10} and continually adapted to a sequence of corruptions with a batch size of 32.
Our method significantly outperforms other baselines in online error $\mathcal{E}(\psi)$, average accuracy $\mathcal{A}(\psi)$, and forward transfer $\mathcal{F}(\psi)$.
}
\label{tab:add-result}
\resizebox{\linewidth}{!}{
\begin{tabular}{l|l|ccccccccccccccc|ccc}
\toprule
\multirow{2}{*}{Level} & \multirow{2}{*}{Method} & \multicolumn{15}{r}{$t$ \Arrow{15.7cm}} & \multirow{2}{*}{$\mathcal{E}(\psi)$ $\downarrow$} &
    \multirow{2}{*}{$\mathcal{A}(\psi) \uparrow$} & \multirow{2}{*}{$\mathcal{F}(\psi) \uparrow$} \\
& & gauss & shot & impul & defoc & glass & motn & zoom & snow & frost & fog & brit & contr & elast & pixel & jpeg &\\
\midrule 
\multirow{6}{*}{4} & ERM & 68.23 & 57.38 & 54.03 & 13.99 & 66.91 & 27.14 & 20.42 & 29.06 & 35.58 & 11.44 & \textbf{9.86} & \textbf{12.11} & 27.59 & 57.57 & 32.91 & 34.95 & 65.05 & - \\
& AdaBN & 37.13 & 32.78 & 32.83 & 15.53 & 51.74 & 22.21 & 19.08 & 27.53 & 25.37 & 15.61 & 11.96 & 15.95 & 26.26 & 24.18 & 32.14 & 26.02 & 73.98 & - \\
& TTT & 36.94 & 24.67 & 30.79 & 20.42 & 40.74 & 25.44 & 17.95 & 27.31 & 18.49 & 14.97 & 12.64 & 15.35 & 22.18 & 19.65 & \textbf{21.58} & 23.27 & 73.09 & \textbf{0.13} \\	
& Tent & 27.58 & \textbf{20.17} & \textbf{23.50} & 14.18 & \textbf{37.28} & 19.16 & 15.77 & 23.20 & 19.59 & 14.39 & 12.77 & 14.18 & 21.87 & \textbf{17.61} & 21.61 & 20.19 & 79.31 & -1.55 \\	
& CoTTA & 26.47 & 23.80 & 26.69 & 18.61 & 43.18 & 25.55 & 24.70 & 28.05 & 26.82 & 25.11 & 17.67 & 31.18 & 36.98 & 31.83 & 35.06 & 28.11 & 63.32 & -12.35 \\
\cmidrule{2-20}
& Ours & \textbf{25.64} & 21.43 & 25.54 & \textbf{12.08} & 40.49 & \textbf{16.27} & \textbf{14.41} & \textbf{21.38} & \textbf{17.36} & \textbf{13.08} & 10.01 & 13.07 & \textbf{20.05} & 17.97 & 22.62 & \textbf{19.43} & \textbf{80.65} & -0.06 \\	
\midrule 
\multirow{6}{*}{3} & ERM & 62.71 & 49.81 & 30.51 & 11.30 & 55.19 & 27.97 & 17.07 & 26.83 & 33.17 & \textbf{10.01} & \textbf{9.62} & \textbf{10.73} & 16.60 & 33.55 & 28.99 & 28.27 & 71.73 & - \\
& AdaBN & 34.24 & 29.07 & 24.69 & 13.14 & 39.59 & 22.09 & 16.97 & 25.50 & 23.81 & 13.80 & 11.94 & 14.39 & 18.48 & 18.28 & 29.93 & 22.39 & 77.61 & - \\
& TTT & 35.07 & 22.45 & 24.02 & 14.71 & 30.25 & 25.08 & 16.67 & 23.46 & 18.01 & 12.64 & 11.19 & 12.86 & 15.42 & 15.00 & \textbf{19.59} & 19.82 & 78.23 & \textbf{1.99} \\	
& Tent & 25.25 & \textbf{17.92} & 18.23 & 11.83 & \textbf{27.75} & 18.52 & 14.49 & 19.64 & 18.37 & 12.86 & 11.84 & 12.89 & 15.55 & 14.90 & 20.27 & 17.35 & 82.41 & -1.11 \\	
& CoTTA & 24.75 & 21.55 & 21.01 & 15.33 & 31.85 & 24.39 & 22.25 & 24.67 & 25.55 & 20.31 & 16.36 & 25.01 & 29.41 & 25.81 & 31.79 & 24.00 & 68.89 & -10.45 \\
\cmidrule{2-20}
& Ours & \textbf{23.47} & 18.85 & \textbf{17.80} & \textbf{9.97} & 28.74 & \textbf{16.51} & \textbf{12.86} & \textbf{18.84} & \textbf{17.06} & 11.18 & 9.76 & 11.48 & \textbf{14.24} & \textbf{13.81} & 20.76 & \textbf{16.35} & \textbf{83.64} & -0.26 \\
\midrule 
\multirow{6}{*}{2} & ERM & 46.90 & 28.94 & 21.62 & \textbf{9.54} & 56.80 & 19.92 & 14.27 & 31.57 & 21.03 & \textbf{9.54} & \textbf{9.21} & \textbf{10.00} & 14.29 & 24.50 & 25.88 & 22.93 & 77.07 & - \\
& AdaBN & 27.66 & 21.47 & 20.29 & 11.95 & 38.96 & 17.88 & 15.02 & 25.55 & 18.75 & 12.61 & 11.76 & 13.24 & 17.00 & 16.30 & 26.68 & 19.67 & 80.33 & - \\
& TTT & 29.04 & 16.88 & 21.81 & 11.52 & 31.61 & 20.59 & 14.72 & 23.47 & 15.46 & 11.06 & 10.29 & 11.25 & 14.65 & 13.73 & 18.51 & 17.64 & 81.74 & \textbf{2.68} \\
& Tent & 21.22 & \textbf{14.21} & 15.33 & 10.49 & \textbf{27.98} & 15.44 & 12.79 & 19.53 & 14.98 & 11.51 & 10.95 & 11.69 & 14.86 & 13.50 & \textbf{17.80} & 15.49 & \textbf{85.76} & -0.42 \\	
& CoTTA & 20.58 & 16.33 & 17.67 & 13.29 & 31.33 & 20.23 & 19.69 & 23.14 & 21.12 & 17.41 & 15.46 & 21.03 & 24.92 & 23.51 & 29.21 & 20.99 & 72.15 & -9.64 \\
\cmidrule{2-20}
& Ours & \textbf{19.48} & 14.43 & \textbf{15.14} & 9.65 & 28.36 & \textbf{13.73} & \textbf{11.37} & \textbf{18.13} & \textbf{13.87} & 10.24 & 9.35 & 10.84 & \textbf{13.13} & \textbf{12.79} & 18.98 & \textbf{14.63} & 85.42 & -0.42 \\	
\midrule 
\multirow{6}{*}{1} & ERM & 27.70 & 19.21 & 15.06 & \textbf{8.80} & 59.08 & 13.64 & 13.44 & 17.35 & 14.61 & \textbf{8.90} & \textbf{8.83} & \textbf{9.00} & 14.37 & 13.32 & 17.98 & 17.42 & 82.58 & - \\
& AdaBN & 20.42 & 17.93 & 15.20 & 11.69 & 38.83 & 14.90 & 14.51 & 17.44 & 16.05 & 11.83 & 11.65 & 12.01 & 17.22 & 14.37 & 20.79 & 16.99 & 83.01 & - \\
& TTT & 23.69 & 14.64 & 17.01 & 9.90 & 33.45 & 15.58 & 14.21 & 16.24 & 11.77 & 9.98 & 9.57 & 9.81 & \textbf{13.31} & 11.72 & 15.18 & 15.07 & 85.45 & \textbf{3.61} \\
& Tent & 16.10 & \textbf{12.54} & 12.37 & 9.70 & \textbf{26.92} & 13.08 & 12.32 & 14.19 & 12.21 & 10.31 & 10.32 & 10.44 & 14.71 & 11.51 & \textbf{14.38} & 13.41 & 86.72 & 0.09 \\	
& CoTTA & 15.95 & 14.74 & 14.41 & 12.07 & 30.40 & 16.58 & 17.82 & 17.47 & 17.72 & 15.41 & 14.63 & 16.67 & 24.28 & 20.19 & 24.07 & 18.16 & 76.26 & -7.72 \\
\cmidrule{2-20}
& Ours & \textbf{14.97} & 12.73 & \textbf{12.12} & 9.45 & 28.98 & \textbf{11.51} & \textbf{11.26} & \textbf{13.22} & \textbf{11.70} & 9.67 & 9.46 & 9.88 & 13.79 & \textbf{11.39} & 14.99 & \textbf{13.01} & \textbf{86.87} & -0.75 \\	
\bottomrule
\end{tabular}
}
\end{table*}

\section{Implementation Details}~\label{app:impl}

We provide the details of network architectures used for the three modalities in our experiments. Following Fig.~\ref{fig:models}, we will describe the design in terms of the feature extractor $\phi$, the main classifier $\theta_m$ and the auxiliary classifier $\theta_a$ for easy interpretation. All experiments are conducted on a single GeForce RTX 2080 Ti GPU with 12G memory.

\subsection{Image}\label{sec:image_a}
\subsubsection{CIFAR-10/100-C}
We use ResNet-26 \cite{he2016deep} as the backbone. The input dimension is $3 \times 32 \times 32$.

$\phi$: The first layer of the model contains a $3 \times 3$ convolution (Conv) layer with 16 output channels. The input is then passed through three ResNet layers, each having four basic blocks. 
Each basic block consists of two Conv $3 \times 3$ layers with the same number of channels. The number of channels for each ResNet layer is \{16, 32, 64\}. 
Each Conv is followed by batch normalization (BN) and a rectifier linear unit (ReLU). 
After all the Conv layers, the network produces a feature map of size $64 \times 8 \times 8$, reduced to $64 \times 1 \times 1$ after pooling. 

$\theta_m$: The feature map is flattened and fed into a fully connected (FC) layer of size 10/100 to generate the logits before applying Softmax for the final prediction.

$\theta_a$: We then append a two-layer FC network with size \{64, 10/100\} that take the logits from $\theta_m$ as the input. 

\subsubsection{Tiny-ImageNet-C}
We use ResNet-34 \cite{he2016deep} as the backbone. The input is resized to $3 \times 224 \times 224$.

$\phi$: The module follows the official PyTorch \cite{pytorchbib} implementation.

$\theta_m$: The feature map is flattened and fed into an FC layer of size 200 to generate the logits before applying the softmax activation function for the final prediction.

$\theta_a$: We then append a two-layer FC network with size \{1024, 200\} that take the logits from $\theta_m$ as the input.

We use the Stochastic Gradient Descent optimizer with Nesterov momentum \cite{Sutskever2013OnTI} and set the batch size to 32 for {\it CIFAR-10/100-C} and 64 for {\it Tiny-ImageNet-C}. During pre-training, the initial learning rate is 0.001 with a linear decay and the number of epochs is 75. During adaptation, the learning rate is set to 0.001.

\subsection{Text}\label{sec:text_a}
We employ four data augmentations:
(1) Synonym replacement: randomly choose words from the sentence that are not stop words and replace each of them with one of their synonyms chosen at random;
(2) Random insertion: find a random synonym of a random word in the sentence that is not a stop word and insert that synonym into a random position in the sentence;
(3) Random swap: randomly choose a pair of words in the sentence and swap their positions;
(4) Random deletion: randomly remove each word in the sentence with a certain probability.
For all augmentations, we use the hyper-parameters from~\cite{wei2019eda}. We use a simple FC network. Raw text samples are first converted into the mSDA \cite{chen2012marginalized} representation. 

$\phi$: The mSDA vector is then fed into two FC layers with the hidden dimensions of 1024 and 512. 

$\theta_m$: The output feature is then passed into an FC layer of size 2 and a softmax layer for binary prediction.

$\theta_a$: Another FC layer of size 2 is appended to $\theta_m$.

We append a BN layer to each hidden layer. We use Adam \cite{kingma2014adam} to optimize the model for 1000 iterations, and the initial learning rate $\eta_{0}=10^{-4}$. We adopt the inverse-decay strategy of DANN \cite{zhang2016understanding}, where the learning rate changes by $\eta_{p}=\frac{\eta_{0}}{(1+\omega p)^{\phi}}, \omega=10$, $\phi=0.75$, and $p$ is the progress ranging from 0 to 1. Similar to image classification, the batch size is 32, simulating the online setting.

\subsection{Speech}\label{sec:speech_a}
In {\it Google Commands} \cite{warden2018speech}, the range of ``amplitude change'' is (0.7,1.1). The maximum scales of ``pitch change'', ``background noise'', and ``stretch'' are 0.2, 0.45, and 0.2, respectively. The maximum shift of ``time shift'' is 8. 
For audio augmentation, we leverage the following three operations:
(1) Gain: multiply audio by a random amplitude factor to lower or raise volume;
(2) High-pass filtering: apply parameterized filter steepness to input audio; 
(3) Impulse response: convolve audio with a randomly selected impulse response.
The mel-spectrogram features of dimension $1 \times 32 \times 32$ are fed into LeNet \cite{Lecun98gradient} as the $1$-channel image. 

$\phi$: The original image is fed into two $5 \times 5$ Conv layers with the channels of 6 and 16, outputing a feature of dimension $16 \times 5 \times 5$.

$\theta_m$: The feature then go through two FC layers with the size of 120 and 84, respectively. The output is a 30-dimensional vector before applying softmax to predict the spoken word.

$\theta_a$: An FC layer of size 30 follows $\theta_m$.

We append a BN layer to each Conv layer. Models are trained using SGD with a learning rate of $0.1$ linearly reduced to $0.001$ for 50 epochs.




\begin{table}
    \centering
    \caption{Average time taken to adapt to each corruption on \textit{CIFAR-10/100-C} and \textit{Tiny-ImageNet-C}.}
    \resizebox{0.85\linewidth}{!}{
    \begin{tabular}{l|c|c|c}
    \toprule
         Methods & CIFAR-10-C & CIFAR-100-C & Tiny-ImageNet-C \\
    \hline
         CoTTA & 24 sec & 36 sec & 175 sec\\
         Ours & \textbf{7 sec} & \textbf{7 sec} & \textbf{19 sec} \\
    \bottomrule
    \end{tabular}
    }
    
    \label{tab:speed}
\end{table}

\section{Additional Results}

We report the results on {\it CIFAR-10-C} in the instantaneously changing setup on severity levels $4$ to $1$ (Tab.~\ref{tab:add-result}). 
We use the same baselines as in Sec.~\ref{sec: experiments} of the main paper: ERM \cite{vapnik1998statistical}, AdaBN \cite{Li2018AdaptiveBN}, TTT \cite{sun2019test}, Tent \cite{wang2021tent}, and CoTTA \cite{Wang2022ContinualTD}.
Overall, our method consistently outperforms others in terms of $\mathcal{E}(\psi)$ and $\mathcal{A}(\psi)$ across the four corruption severity levels. Notably, CoTTA displays a substantial loss in both $\mathcal{F}(\psi)$ possibly due to two reasons. First, the error accumulation due to insufficient statistics (\ie small batch size) deteriorates the model performance over time, leading to the incapability to transfer past domains' knowledge to the current. Second, by updating all the parameters, CoTTA suffers from catastrophic forgetting since the model overfits to the data stream. 
Experiments show that only updating a portion of the learnable parameters helps retain past knowledge, thus avoiding forgetting.

\section{Inference Time}
Different from CoTTA which updates all the parameters of the feature extractor, our method only updates BN parameters.
Therefore, our methods are an order of magnitude faster than CoTTA (Tab.~\ref{tab:speed}).


\bibliographystyle{ACM-Reference-Format}
\bibliography{sample-base}


\begin{thebibliography}{52}


\ifx \showCODEN    \undefined \def \showCODEN     #1{\unskip}     \fi
\ifx \showDOI      \undefined \def \showDOI       #1{#1}\fi
\ifx \showISBNx    \undefined \def \showISBNx     #1{\unskip}     \fi
\ifx \showISBNxiii \undefined \def \showISBNxiii  #1{\unskip}     \fi
\ifx \showISSN     \undefined \def \showISSN      #1{\unskip}     \fi
\ifx \showLCCN     \undefined \def \showLCCN      #1{\unskip}     \fi
\ifx \shownote     \undefined \def \shownote      #1{#1}          \fi
\ifx \showarticletitle \undefined \def \showarticletitle #1{#1}   \fi
\ifx \showURL      \undefined \def \showURL       {\relax}        \fi
\providecommand\bibfield[2]{#2}
\providecommand\bibinfo[2]{#2}
\providecommand\natexlab[1]{#1}
\providecommand\showeprint[2][]{arXiv:#2}

\bibitem[Aljundi et~al\mbox{.}(2017)]%
        {aljundi2017expert}
\bibfield{author}{\bibinfo{person}{Rahaf Aljundi}, \bibinfo{person}{Punarjay Chakravarty}, {and} \bibinfo{person}{Tinne Tuytelaars}.} \bibinfo{year}{2017}\natexlab{}.
\newblock \showarticletitle{Expert gate: Lifelong learning with a network of experts}. In \bibinfo{booktitle}{\emph{Proceedings of the IEEE Conference on Computer Vision and Pattern Recognition}}. \bibinfo{pages}{3366--3375}.
\newblock


\bibitem[Balaji et~al\mbox{.}(2018)]%
        {balaji2018metareg}
\bibfield{author}{\bibinfo{person}{Yogesh Balaji}, \bibinfo{person}{Swami Sankaranarayanan}, {and} \bibinfo{person}{Rama Chellappa}.} \bibinfo{year}{2018}\natexlab{}.
\newblock \showarticletitle{Metareg: Towards domain generalization using meta-regularization}. In \bibinfo{booktitle}{\emph{NeurIPS}}. \bibinfo{pages}{998--1008}.
\newblock


\bibitem[Bartler et~al\mbox{.}(2021)]%
        {bartler2022mt3}
\bibfield{author}{\bibinfo{person}{Alexander Bartler}, \bibinfo{person}{Andreas B{\"u}hler}, \bibinfo{person}{Felix Wiewel}, \bibinfo{person}{Mario D{\"o}bler}, {and} \bibinfo{person}{Binh Yang}.} \bibinfo{year}{2021}\natexlab{}.
\newblock \showarticletitle{MT3: Meta Test-Time Training for Self-Supervised Test-Time Adaption}. In \bibinfo{booktitle}{\emph{International Conference on Artificial Intelligence and Statistics}}.
\newblock
\urldef\tempurl%
\url{https://api.semanticscholar.org/CorpusID:232417890}
\showURL{%
\tempurl}


\bibitem[Ben-David et~al\mbox{.}(2006)]%
        {ben2006analysis}
\bibfield{author}{\bibinfo{person}{Shai Ben-David}, \bibinfo{person}{John Blitzer}, \bibinfo{person}{Koby Crammer}, {and} \bibinfo{person}{Fernando Pereira}.} \bibinfo{year}{2006}\natexlab{}.
\newblock \showarticletitle{Analysis of representations for domain adaptation}.
\newblock \bibinfo{journal}{\emph{Advances in neural information processing systems}}  \bibinfo{volume}{19} (\bibinfo{year}{2006}).
\newblock


\bibitem[Chaudhry et~al\mbox{.}(2018)]%
        {chaudhry2018efficient}
\bibfield{author}{\bibinfo{person}{Arslan Chaudhry}, \bibinfo{person}{Marc’Aurelio Ranzato}, \bibinfo{person}{Marcus Rohrbach}, {and} \bibinfo{person}{Mohamed Elhoseiny}.} \bibinfo{year}{2018}\natexlab{}.
\newblock \showarticletitle{Efficient Lifelong Learning with A-GEM}. In \bibinfo{booktitle}{\emph{International Conference on Learning Representations}}.
\newblock


\bibitem[Chen et~al\mbox{.}(2012)]%
        {chen2012marginalized}
\bibfield{author}{\bibinfo{person}{Minmin Chen}, \bibinfo{person}{Zhixiang Xu}, \bibinfo{person}{Kilian~Q Weinberger}, {and} \bibinfo{person}{Fei Sha}.} \bibinfo{year}{2012}\natexlab{}.
\newblock \showarticletitle{Marginalized denoising autoencoders for domain adaptation}. In \bibinfo{booktitle}{\emph{ICML}}. \bibinfo{pages}{1627--1634}.
\newblock


\bibitem[De~Lange et~al\mbox{.}(2021)]%
        {de2021continual}
\bibfield{author}{\bibinfo{person}{Matthias De~Lange}, \bibinfo{person}{Rahaf Aljundi}, \bibinfo{person}{Marc Masana}, \bibinfo{person}{Sarah Parisot}, \bibinfo{person}{Xu Jia}, \bibinfo{person}{Ale{\v{s}} Leonardis}, \bibinfo{person}{Gregory Slabaugh}, {and} \bibinfo{person}{Tinne Tuytelaars}.} \bibinfo{year}{2021}\natexlab{}.
\newblock \showarticletitle{A continual learning survey: Defying forgetting in classification tasks}.
\newblock \bibinfo{journal}{\emph{IEEE transactions on pattern analysis and machine intelligence}} \bibinfo{volume}{44}, \bibinfo{number}{7} (\bibinfo{year}{2021}), \bibinfo{pages}{3366--3385}.
\newblock


\bibitem[Dou et~al\mbox{.}(2019)]%
        {dou2019domain}
\bibfield{author}{\bibinfo{person}{Qi Dou}, \bibinfo{person}{Daniel~Coelho de Castro}, \bibinfo{person}{Konstantinos Kamnitsas}, {and} \bibinfo{person}{Ben Glocker}.} \bibinfo{year}{2019}\natexlab{}.
\newblock \showarticletitle{Domain generalization via model-agnostic learning of semantic features}. In \bibinfo{booktitle}{\emph{NeurIPS}}.
\newblock


\bibitem[Finn et~al\mbox{.}(2017)]%
        {finn2017model}
\bibfield{author}{\bibinfo{person}{Chelsea Finn}, \bibinfo{person}{Pieter Abbeel}, {and} \bibinfo{person}{Sergey Levine}.} \bibinfo{year}{2017}\natexlab{}.
\newblock \showarticletitle{Model-agnostic meta-learning for fast adaptation of deep networks}. In \bibinfo{booktitle}{\emph{ICML}}.
\newblock


\bibitem[Ganin and Lempitsky(2015)]%
        {ganin2015unsupervised}
\bibfield{author}{\bibinfo{person}{Yaroslav Ganin} {and} \bibinfo{person}{Victor Lempitsky}.} \bibinfo{year}{2015}\natexlab{}.
\newblock \showarticletitle{{Unsupervised Domain Adaptation by Backpropagation}}. In \bibinfo{booktitle}{\emph{ICML}}. \bibinfo{pages}{1180--1189}.
\newblock


\bibitem[Goodfellow et~al\mbox{.}(2015)]%
        {goodfellow2014adv}
\bibfield{author}{\bibinfo{person}{Ian~J Goodfellow}, \bibinfo{person}{Jonathon Shlens}, {and} \bibinfo{person}{Christian Szegedy}.} \bibinfo{year}{2015}\natexlab{}.
\newblock \showarticletitle{Explaining and harnessing adversarial examples}. In \bibinfo{booktitle}{\emph{International Conference on Learning Representations}}.
\newblock


\bibitem[He et~al\mbox{.}(2016)]%
        {he2016deep}
\bibfield{author}{\bibinfo{person}{Kaiming He}, \bibinfo{person}{Xiangyu Zhang}, \bibinfo{person}{Shaoqing Ren}, {and} \bibinfo{person}{Jian Sun}.} \bibinfo{year}{2016}\natexlab{}.
\newblock \showarticletitle{{Deep Residual Learning for Image Recognition}}. In \bibinfo{booktitle}{\emph{CVPR}}. \bibinfo{pages}{770--778}.
\newblock


\bibitem[Hendrycks and Dietterich(2019)]%
        {hendrycks2019benchmarking}
\bibfield{author}{\bibinfo{person}{Dan Hendrycks} {and} \bibinfo{person}{Thomas Dietterich}.} \bibinfo{year}{2019}\natexlab{}.
\newblock \showarticletitle{Benchmarking neural network robustness to common corruptions and perturbations}.
\newblock \bibinfo{journal}{\emph{ICLR}} (\bibinfo{year}{2019}).
\newblock


\bibitem[Hendrycks et~al\mbox{.}(2020)]%
        {hendrycks2020augmix}
\bibfield{author}{\bibinfo{person}{Dan Hendrycks}, \bibinfo{person}{Norman Mu}, \bibinfo{person}{Ekin~D. Cubuk}, \bibinfo{person}{Barret Zoph}, \bibinfo{person}{Justin Gilmer}, {and} \bibinfo{person}{Balaji Lakshminarayanan}.} \bibinfo{year}{2020}\natexlab{}.
\newblock \showarticletitle{{AugMix}: A Simple Data Processing Method to Improve Robustness and Uncertainty}.
\newblock \bibinfo{journal}{\emph{ICLR}} (\bibinfo{year}{2020}).
\newblock


\bibitem[Hoffman et~al\mbox{.}(2014)]%
        {hoffman2014continuous}
\bibfield{author}{\bibinfo{person}{Judy Hoffman}, \bibinfo{person}{Trevor Darrell}, {and} \bibinfo{person}{Kate Saenko}.} \bibinfo{year}{2014}\natexlab{}.
\newblock \showarticletitle{Continuous Manifold Based Adaptation for Evolving Visual Domains}.
\newblock \bibinfo{journal}{\emph{2014 IEEE Conference on Computer Vision and Pattern Recognition}} (\bibinfo{year}{2014}), \bibinfo{pages}{867--874}.
\newblock
\urldef\tempurl%
\url{https://api.semanticscholar.org/CorpusID:10105727}
\showURL{%
\tempurl}


\bibitem[Huang et~al\mbox{.}(2022)]%
        {huang2022lifelong}
\bibfield{author}{\bibinfo{person}{Zhipeng Huang}, \bibinfo{person}{Zhizheng Zhang}, \bibinfo{person}{Cuiling Lan}, \bibinfo{person}{Wenjun Zeng}, \bibinfo{person}{Peng Chu}, \bibinfo{person}{Quanzeng You}, \bibinfo{person}{Jiang Wang}, \bibinfo{person}{Zicheng Liu}, {and} \bibinfo{person}{Zheng-jun Zha}.} \bibinfo{year}{2022}\natexlab{}.
\newblock \showarticletitle{Lifelong unsupervised domain adaptive person re-identification with coordinated anti-forgetting and adaptation}. In \bibinfo{booktitle}{\emph{Proceedings of the IEEE/CVF Conference on Computer Vision and Pattern Recognition}}. \bibinfo{pages}{14288--14297}.
\newblock


\bibitem[Kingma and Ba(2014)]%
        {kingma2014adam}
\bibfield{author}{\bibinfo{person}{Diederik~P. Kingma} {and} \bibinfo{person}{Jimmy Ba}.} \bibinfo{year}{2014}\natexlab{}.
\newblock \showarticletitle{{Adam: A Method for Stochastic Optimization}}. In \bibinfo{booktitle}{\emph{arXiv:1412.6980 [cs.LG]}}.
\newblock


\bibitem[Kirkpatrick et~al\mbox{.}(2017)]%
        {kirkpatrick2017overcoming}
\bibfield{author}{\bibinfo{person}{James Kirkpatrick}, \bibinfo{person}{Razvan Pascanu}, \bibinfo{person}{Neil Rabinowitz}, \bibinfo{person}{Joel Veness}, \bibinfo{person}{Guillaume Desjardins}, \bibinfo{person}{Andrei~A Rusu}, \bibinfo{person}{Kieran Milan}, \bibinfo{person}{John Quan}, \bibinfo{person}{Tiago Ramalho}, \bibinfo{person}{Agnieszka Grabska-Barwinska}, {et~al\mbox{.}}} \bibinfo{year}{2017}\natexlab{}.
\newblock \showarticletitle{Overcoming catastrophic forgetting in neural networks}.
\newblock \bibinfo{journal}{\emph{Proceedings of the national academy of sciences}} \bibinfo{volume}{114}, \bibinfo{number}{13} (\bibinfo{year}{2017}), \bibinfo{pages}{3521--3526}.
\newblock


\bibitem[Krizhevsky et~al\mbox{.}(2009)]%
        {krizhevsky2009learning}
\bibfield{author}{\bibinfo{person}{Alex Krizhevsky}, \bibinfo{person}{Geoffrey Hinton}, {et~al\mbox{.}}} \bibinfo{year}{2009}\natexlab{}.
\newblock \showarticletitle{Learning multiple layers of features from tiny images}.
\newblock  (\bibinfo{year}{2009}).
\newblock


\bibitem[Lao et~al\mbox{.}(2021)]%
        {lao2021two}
\bibfield{author}{\bibinfo{person}{Qicheng Lao}, \bibinfo{person}{Xiangxi Jiang}, \bibinfo{person}{Mohammad Havaei}, {and} \bibinfo{person}{Yoshua Bengio}.} \bibinfo{year}{2021}\natexlab{}.
\newblock \showarticletitle{A Two-Stream Continual Learning System With Variational Domain-Agnostic Feature Replay}.
\newblock \bibinfo{journal}{\emph{IEEE Transactions on Neural Networks and Learning Systems}}  \bibinfo{volume}{33} (\bibinfo{year}{2021}), \bibinfo{pages}{4466--4478}.
\newblock
\urldef\tempurl%
\url{https://api.semanticscholar.org/CorpusID:232113812}
\showURL{%
\tempurl}


\bibitem[Le and Yang(2015)]%
        {Le2015TinyIV}
\bibfield{author}{\bibinfo{person}{Y. Le} {and} \bibinfo{person}{X. Yang}.} \bibinfo{year}{2015}\natexlab{}.
\newblock \showarticletitle{Tiny ImageNet Visual Recognition Challenge}.
\newblock


\bibitem[Lecun et~al\mbox{.}(1998)]%
        {Lecun98gradient}
\bibfield{author}{\bibinfo{person}{Yann Lecun} {et~al\mbox{.}}} \bibinfo{year}{1998}\natexlab{}.
\newblock \showarticletitle{{Gradient-Based Learning Applied to Document Recognition}}.
\newblock \bibinfo{journal}{\emph{Proc. IEEE}} \bibinfo{volume}{86}, \bibinfo{number}{11} (\bibinfo{year}{1998}).
\newblock


\bibitem[Li et~al\mbox{.}(2017)]%
        {li2018learning}
\bibfield{author}{\bibinfo{person}{Da Li}, \bibinfo{person}{Yongxin Yang}, \bibinfo{person}{Yi-Zhe Song}, {and} \bibinfo{person}{Timothy~M. Hospedales}.} \bibinfo{year}{2017}\natexlab{}.
\newblock \showarticletitle{Learning to Generalize: Meta-Learning for Domain Generalization}. In \bibinfo{booktitle}{\emph{AAAI Conference on Artificial Intelligence}}.
\newblock
\urldef\tempurl%
\url{https://api.semanticscholar.org/CorpusID:1883787}
\showURL{%
\tempurl}


\bibitem[Li et~al\mbox{.}(2018)]%
        {Li2018AdaptiveBN}
\bibfield{author}{\bibinfo{person}{Yanghao Li}, \bibinfo{person}{Naiyan Wang}, \bibinfo{person}{Jianping Shi}, \bibinfo{person}{Xiaodi Hou}, {and} \bibinfo{person}{Jiaying Liu}.} \bibinfo{year}{2018}\natexlab{}.
\newblock \showarticletitle{Adaptive Batch Normalization for practical domain adaptation}.
\newblock \bibinfo{journal}{\emph{Pattern Recognit.}}  \bibinfo{volume}{80} (\bibinfo{year}{2018}), \bibinfo{pages}{109--117}.
\newblock


\bibitem[Li et~al\mbox{.}(2016)]%
        {li2016revisiting}
\bibfield{author}{\bibinfo{person}{Yanghao Li}, \bibinfo{person}{Naiyan Wang}, \bibinfo{person}{Jianping Shi}, \bibinfo{person}{Jiaying Liu}, {and} \bibinfo{person}{Xiaodi Hou}.} \bibinfo{year}{2016}\natexlab{}.
\newblock \showarticletitle{Revisiting Batch Normalization For Practical Domain Adaptation}.
\newblock \bibinfo{journal}{\emph{ArXiv}}  \bibinfo{volume}{abs/1603.04779} (\bibinfo{year}{2016}).
\newblock
\urldef\tempurl%
\url{https://api.semanticscholar.org/CorpusID:5069968}
\showURL{%
\tempurl}


\bibitem[Li and Hoiem(2017)]%
        {li2017learning}
\bibfield{author}{\bibinfo{person}{Zhizhong Li} {and} \bibinfo{person}{Derek Hoiem}.} \bibinfo{year}{2017}\natexlab{}.
\newblock \showarticletitle{Learning without forgetting}.
\newblock \bibinfo{journal}{\emph{IEEE transactions on pattern analysis and machine intelligence}} \bibinfo{volume}{40}, \bibinfo{number}{12} (\bibinfo{year}{2017}), \bibinfo{pages}{2935--2947}.
\newblock


\bibitem[Liang et~al\mbox{.}(2020)]%
        {liang2020we}
\bibfield{author}{\bibinfo{person}{Jian Liang}, \bibinfo{person}{Dapeng Hu}, {and} \bibinfo{person}{Jiashi Feng}.} \bibinfo{year}{2020}\natexlab{}.
\newblock \showarticletitle{Do we really need to access the source data? source hypothesis transfer for unsupervised domain adaptation}. In \bibinfo{booktitle}{\emph{International Conference on Machine Learning}}. PMLR, \bibinfo{pages}{6028--6039}.
\newblock


\bibitem[Liu et~al\mbox{.}(2020)]%
        {liu2020learning}
\bibfield{author}{\bibinfo{person}{Hong Liu}, \bibinfo{person}{Mingsheng Long}, \bibinfo{person}{Jianmin Wang}, {and} \bibinfo{person}{Yu Wang}.} \bibinfo{year}{2020}\natexlab{}.
\newblock \showarticletitle{Learning to Adapt to Evolving Domains}. In \bibinfo{booktitle}{\emph{Neural Information Processing Systems}}.
\newblock
\urldef\tempurl%
\url{https://api.semanticscholar.org/CorpusID:227275334}
\showURL{%
\tempurl}


\bibitem[Liu et~al\mbox{.}(2021)]%
        {liu2021ttt}
\bibfield{author}{\bibinfo{person}{Yuejiang Liu}, \bibinfo{person}{Parth Kothari}, \bibinfo{person}{Bastien van Delft}, \bibinfo{person}{Baptiste Bellot-Gurlet}, \bibinfo{person}{Taylor Mordan}, {and} \bibinfo{person}{Alexandre Alahi}.} \bibinfo{year}{2021}\natexlab{}.
\newblock \showarticletitle{Ttt++: When does self-supervised test-time training fail or thrive?}
\newblock \bibinfo{journal}{\emph{NeurIPS}} (\bibinfo{year}{2021}).
\newblock


\bibitem[Mallya and Lazebnik(2018)]%
        {mallya2018packnet}
\bibfield{author}{\bibinfo{person}{Arun Mallya} {and} \bibinfo{person}{Svetlana Lazebnik}.} \bibinfo{year}{2018}\natexlab{}.
\newblock \showarticletitle{Packnet: Adding multiple tasks to a single network by iterative pruning}. In \bibinfo{booktitle}{\emph{Proceedings of the IEEE conference on Computer Vision and Pattern Recognition}}. \bibinfo{pages}{7765--7773}.
\newblock


\bibitem[McCloskey and Cohen(1989)]%
        {mccloskey1989catastrophic}
\bibfield{author}{\bibinfo{person}{Michael McCloskey} {and} \bibinfo{person}{Neal~J Cohen}.} \bibinfo{year}{1989}\natexlab{}.
\newblock \showarticletitle{Catastrophic interference in connectionist networks: The sequential learning problem}.
\newblock In \bibinfo{booktitle}{\emph{Psychology of learning and motivation}}. Vol.~\bibinfo{volume}{24}. \bibinfo{publisher}{Elsevier}, \bibinfo{pages}{109--165}.
\newblock


\bibitem[Niu et~al\mbox{.}(2022)]%
        {niu2022efficient}
\bibfield{author}{\bibinfo{person}{Shuaicheng Niu}, \bibinfo{person}{Jiaxiang Wu}, \bibinfo{person}{Yifan Zhang}, \bibinfo{person}{Yaofo Chen}, \bibinfo{person}{Shijian Zheng}, \bibinfo{person}{Peilin Zhao}, {and} \bibinfo{person}{Mingkui Tan}.} \bibinfo{year}{2022}\natexlab{}.
\newblock \showarticletitle{Efficient test-time model adaptation without forgetting}. In \bibinfo{booktitle}{\emph{International conference on machine learning}}. PMLR, \bibinfo{pages}{16888--16905}.
\newblock


\bibitem[Paszke et~al\mbox{.}(2019)]%
        {pytorchbib}
\bibfield{author}{\bibinfo{person}{Adam Paszke}, \bibinfo{person}{Sam Gross}, \bibinfo{person}{Francisco Massa}, \bibinfo{person}{Adam Lerer}, \bibinfo{person}{James Bradbury}, \bibinfo{person}{Gregory Chanan}, \bibinfo{person}{Trevor Killeen}, \bibinfo{person}{Zeming Lin}, \bibinfo{person}{Natalia Gimelshein}, \bibinfo{person}{Luca Antiga}, \bibinfo{person}{Alban Desmaison}, \bibinfo{person}{Andreas Kopf}, \bibinfo{person}{Edward Yang}, \bibinfo{person}{Zachary DeVito}, \bibinfo{person}{Martin Raison}, \bibinfo{person}{Alykhan Tejani}, \bibinfo{person}{Sasank Chilamkurthy}, \bibinfo{person}{Benoit Steiner}, \bibinfo{person}{Lu Fang}, \bibinfo{person}{Junjie Bai}, {and} \bibinfo{person}{Soumith Chintala}.} \bibinfo{year}{2019}\natexlab{}.
\newblock \showarticletitle{PyTorch: An Imperative Style, High-Performance Deep Learning Library}.
\newblock In \bibinfo{booktitle}{\emph{Advances in Neural Information Processing Systems 32}}. \bibinfo{publisher}{Curran Associates, Inc.}, \bibinfo{pages}{8024--8035}.
\newblock
\urldef\tempurl%
\url{http://papers.neurips.cc/paper/9015-pytorch-an-imperative-style-high-performance-deep-learning-library.pdf}
\showURL{%
\tempurl}


\bibitem[Prabhudesai et~al\mbox{.}(2023)]%
        {prabhudesai2023test}
\bibfield{author}{\bibinfo{person}{Mihir Prabhudesai}, \bibinfo{person}{Anirudh Goyal}, \bibinfo{person}{Sujoy Paul}, \bibinfo{person}{Sjoerd Van~Steenkiste}, \bibinfo{person}{Mehdi~SM Sajjadi}, \bibinfo{person}{Gaurav Aggarwal}, \bibinfo{person}{Thomas Kipf}, \bibinfo{person}{Deepak Pathak}, {and} \bibinfo{person}{Katerina Fragkiadaki}.} \bibinfo{year}{2023}\natexlab{}.
\newblock \showarticletitle{Test-time adaptation with slot-centric models}. In \bibinfo{booktitle}{\emph{International Conference on Machine Learning}}. PMLR, \bibinfo{pages}{28151--28166}.
\newblock


\bibitem[Rebuffi et~al\mbox{.}(2017)]%
        {rebuffi2017icarl}
\bibfield{author}{\bibinfo{person}{Sylvestre-Alvise Rebuffi}, \bibinfo{person}{Alexander Kolesnikov}, \bibinfo{person}{Georg Sperl}, {and} \bibinfo{person}{Christoph~H Lampert}.} \bibinfo{year}{2017}\natexlab{}.
\newblock \showarticletitle{icarl: Incremental classifier and representation learning}. In \bibinfo{booktitle}{\emph{Proceedings of the IEEE conference on Computer Vision and Pattern Recognition}}. \bibinfo{pages}{2001--2010}.
\newblock


\bibitem[Rostami(2021)]%
        {rostami2021lifelong}
\bibfield{author}{\bibinfo{person}{Mohammad Rostami}.} \bibinfo{year}{2021}\natexlab{}.
\newblock \showarticletitle{Lifelong domain adaptation via consolidated internal distribution}.
\newblock \bibinfo{journal}{\emph{Advances in Neural Information Processing Systems}}  \bibinfo{volume}{34} (\bibinfo{year}{2021}), \bibinfo{pages}{11172--11183}.
\newblock


\bibitem[Schmidhuber(1987)]%
        {schmidhuber1987evolutionary}
\bibfield{author}{\bibinfo{person}{J{\"u}rgen Schmidhuber}.} \bibinfo{year}{1987}\natexlab{}.
\newblock \emph{\bibinfo{title}{Evolutionary principles in self-referential learning}}.
\newblock \bibinfo{thesistype}{Ph.\,D. Dissertation}. \bibinfo{school}{Technische Universit{\"a}t M{\"u}nchen}.
\newblock


\bibitem[Schneider et~al\mbox{.}(2020)]%
        {schneider2020improving}
\bibfield{author}{\bibinfo{person}{Steffen Schneider}, \bibinfo{person}{Evgenia Rusak}, \bibinfo{person}{Luisa Eck}, \bibinfo{person}{Oliver Bringmann}, \bibinfo{person}{Wieland Brendel}, {and} \bibinfo{person}{Matthias Bethge}.} \bibinfo{year}{2020}\natexlab{}.
\newblock \showarticletitle{Improving robustness against common corruptions by covariate shift adaptation}.
\newblock \bibinfo{journal}{\emph{ArXiv}}  \bibinfo{volume}{abs/2006.16971} (\bibinfo{year}{2020}).
\newblock
\urldef\tempurl%
\url{https://api.semanticscholar.org/CorpusID:220266097}
\showURL{%
\tempurl}


\bibitem[Su et~al\mbox{.}(2020)]%
        {su2020gradient}
\bibfield{author}{\bibinfo{person}{Peng Su}, \bibinfo{person}{Shixiang Tang}, \bibinfo{person}{Peng Gao}, \bibinfo{person}{Di Qiu}, \bibinfo{person}{Ni Zhao}, {and} \bibinfo{person}{Xiaogang Wang}.} \bibinfo{year}{2020}\natexlab{}.
\newblock \showarticletitle{Gradient Regularized Contrastive Learning for Continual Domain Adaptation}. In \bibinfo{booktitle}{\emph{AAAI Conference on Artificial Intelligence}}.
\newblock
\urldef\tempurl%
\url{https://api.semanticscholar.org/CorpusID:220793718}
\showURL{%
\tempurl}


\bibitem[Sun et~al\mbox{.}(2020)]%
        {sun2019test}
\bibfield{author}{\bibinfo{person}{Yu Sun}, \bibinfo{person}{Xiaolong Wang}, \bibinfo{person}{Zhuang Liu}, \bibinfo{person}{John Miller}, \bibinfo{person}{Alexei Efros}, {and} \bibinfo{person}{Moritz Hardt}.} \bibinfo{year}{2020}\natexlab{}.
\newblock \showarticletitle{Test-Time Training with Self-Supervision for Generalization under Distribution Shifts}. In \bibinfo{booktitle}{\emph{Proceedings of the 37th International Conference on Machine Learning}} \emph{(\bibinfo{series}{Proceedings of Machine Learning Research}, Vol.~\bibinfo{volume}{119})}, \bibfield{editor}{\bibinfo{person}{Hal~Daumé III} {and} \bibinfo{person}{Aarti Singh}} (Eds.). \bibinfo{publisher}{PMLR}, \bibinfo{pages}{9229--9248}.
\newblock
\urldef\tempurl%
\url{https://proceedings.mlr.press/v119/sun20b.html}
\showURL{%
\tempurl}


\bibitem[Sutskever et~al\mbox{.}(2013)]%
        {Sutskever2013OnTI}
\bibfield{author}{\bibinfo{person}{Ilya Sutskever}, \bibinfo{person}{James Martens}, \bibinfo{person}{George~E. Dahl}, {and} \bibinfo{person}{Geoffrey~E. Hinton}.} \bibinfo{year}{2013}\natexlab{}.
\newblock \showarticletitle{On the importance of initialization and momentum in deep learning}. In \bibinfo{booktitle}{\emph{ICML}}.
\newblock


\bibitem[Vapnik(1998)]%
        {vapnik1998statistical}
\bibfield{author}{\bibinfo{person}{Vladimir Vapnik}.} \bibinfo{year}{1998}\natexlab{}.
\newblock \bibinfo{title}{Statistical learning theory}.
\newblock
\newblock


\bibitem[V{\'e}niat et~al\mbox{.}(2021)]%
        {Vniat2021EfficientCL}
\bibfield{author}{\bibinfo{person}{Tom V{\'e}niat}, \bibinfo{person}{Ludovic Denoyer}, {and} \bibinfo{person}{Marc'Aurelio Ranzato}.} \bibinfo{year}{2021}\natexlab{}.
\newblock \showarticletitle{Efficient Continual Learning with Modular Networks and Task-Driven Priors}.
\newblock \bibinfo{journal}{\emph{ArXiv}}  \bibinfo{volume}{abs/2012.12631} (\bibinfo{year}{2021}).
\newblock


\bibitem[Volpi et~al\mbox{.}(2021)]%
        {volpi2021continual}
\bibfield{author}{\bibinfo{person}{Riccardo Volpi}, \bibinfo{person}{Diane Larlus}, {and} \bibinfo{person}{Gr{\'e}gory Rogez}.} \bibinfo{year}{2021}\natexlab{}.
\newblock \showarticletitle{Continual adaptation of visual representations via domain randomization and meta-learning}. In \bibinfo{booktitle}{\emph{CVPR}}.
\newblock


\bibitem[Wang et~al\mbox{.}(2021)]%
        {wang2021tent}
\bibfield{author}{\bibinfo{person}{Dequan Wang}, \bibinfo{person}{Evan Shelhamer}, \bibinfo{person}{Shaoteng Liu}, \bibinfo{person}{Bruno~A. Olshausen}, {and} \bibinfo{person}{Trevor Darrell}.} \bibinfo{year}{2021}\natexlab{}.
\newblock \showarticletitle{Tent: Fully Test-Time Adaptation by Entropy Minimization}. In \bibinfo{booktitle}{\emph{International Conference on Learning Representations}}.
\newblock
\urldef\tempurl%
\url{https://api.semanticscholar.org/CorpusID:232278031}
\showURL{%
\tempurl}


\bibitem[Wang et~al\mbox{.}(2022)]%
        {Wang2022ContinualTD}
\bibfield{author}{\bibinfo{person}{Qin Wang}, \bibinfo{person}{Olga Fink}, \bibinfo{person}{Luc Van~Gool}, {and} \bibinfo{person}{Dengxin Dai}.} \bibinfo{year}{2022}\natexlab{}.
\newblock \showarticletitle{Continual test-time domain adaptation}. In \bibinfo{booktitle}{\emph{CVPR}}.
\newblock


\bibitem[Warden(2018)]%
        {warden2018speech}
\bibfield{author}{\bibinfo{person}{Pete Warden}.} \bibinfo{year}{2018}\natexlab{}.
\newblock \showarticletitle{Speech commands: A dataset for limited-vocabulary speech recognition}.
\newblock \bibinfo{journal}{\emph{arXiv preprint arXiv:1804.03209}} (\bibinfo{year}{2018}).
\newblock


\bibitem[Wei and Zou(2019)]%
        {wei2019eda}
\bibfield{author}{\bibinfo{person}{Jason Wei} {and} \bibinfo{person}{Kai Zou}.} \bibinfo{year}{2019}\natexlab{}.
\newblock \showarticletitle{EDA: Easy Data Augmentation Techniques for Boosting Performance on Text Classification Tasks}. In \bibinfo{booktitle}{\emph{Proceedings of the 2019 Conference on Empirical Methods in Natural Language Processing and the 9th International Joint Conference on Natural Language Processing (EMNLP-IJCNLP)}}. \bibinfo{pages}{6382--6388}.
\newblock


\bibitem[Wulfmeier et~al\mbox{.}(2018)]%
        {wulfmeier2018incremental}
\bibfield{author}{\bibinfo{person}{Markus Wulfmeier}, \bibinfo{person}{Alex Bewley}, {and} \bibinfo{person}{Ingmar Posner}.} \bibinfo{year}{2018}\natexlab{}.
\newblock \showarticletitle{Incremental adversarial domain adaptation for continually changing environments}. In \bibinfo{booktitle}{\emph{2018 IEEE International conference on robotics and automation (ICRA)}}. IEEE, \bibinfo{pages}{4489--4495}.
\newblock


\bibitem[Zagoruyko and Komodakis(2016)]%
        {Zagoruyko2016WideRN}
\bibfield{author}{\bibinfo{person}{Sergey Zagoruyko} {and} \bibinfo{person}{Nikos Komodakis}.} \bibinfo{year}{2016}\natexlab{}.
\newblock \showarticletitle{Wide Residual Networks}.
\newblock \bibinfo{journal}{\emph{ArXiv}}  \bibinfo{volume}{abs/1605.07146} (\bibinfo{year}{2016}).
\newblock
\urldef\tempurl%
\url{https://api.semanticscholar.org/CorpusID:15276198}
\showURL{%
\tempurl}


\bibitem[Zhang et~al\mbox{.}(2017)]%
        {zhang2016understanding}
\bibfield{author}{\bibinfo{person}{Chiyuan Zhang}, \bibinfo{person}{Samy Bengio}, \bibinfo{person}{Moritz Hardt}, \bibinfo{person}{Benjamin Recht}, {and} \bibinfo{person}{Oriol Vinyals}.} \bibinfo{year}{2017}\natexlab{}.
\newblock \showarticletitle{Understanding deep learning requires rethinking generalization}. In \bibinfo{booktitle}{\emph{ICLR}}.
\newblock


\bibitem[Zhao et~al\mbox{.}(2023)]%
        {zhao2023pitfalls}
\bibfield{author}{\bibinfo{person}{Hao Zhao}, \bibinfo{person}{Yuejiang Liu}, \bibinfo{person}{Alexandre Alahi}, {and} \bibinfo{person}{Tao Lin}.} \bibinfo{year}{2023}\natexlab{}.
\newblock \bibinfo{title}{On Pitfalls of Test-Time Adaptation}.
\newblock
\newblock
\showeprint[arxiv]{2306.03536}~[cs.LG]


\end{thebibliography}


\end{document}